\documentclass[sigconf]{acmart}

\AtBeginDocument{%
  \providecommand\BibTeX{{%
    \normalfont B\kern-0.5em{\scshape i\kern-0.25em b}\kern-0.8em\TeX}}}

\copyrightyear{2021} 
\acmYear{2021} 
\setcopyright{acmlicensed}\acmConference[MM '21]{Proceedings of the 29th
ACM International Conference on Multimedia}{October 20--24, 2021}{Virtual
Event, China}
\acmBooktitle{Proceedings of the 29th ACM International Conference on
Multimedia (MM '21), October 20--24, 2021, Virtual Event, China}
\acmPrice{15.00}
\acmDOI{10.1145/3474085.3475298}
\acmISBN{978-1-4503-8651-7/21/10}

\acmSubmissionID{719}

\usepackage{graphicx}
\usepackage{multirow}
\usepackage{enumitem}
\usepackage{indentfirst}
\usepackage{lipsum}

\newcommand{\eg}{\textit{e.g.,\ }}
\newcommand{\ie}{\textit{i.e.,\ }}
\newcommand{\et}{\textit{et al.\ }}
\newcommand{\ftone}{\textcolor[rgb]{0,0,0}}
\newcommand{\jcone}{\textcolor[rgb]{0,0,0}}
\newcommand{\fttwo}{\textcolor[rgb]{0.0,0,0.0}}
\newcommand{\jctwo}{\textcolor[rgb]{0,0.0,0.0}}
\newcommand{\ftthree}{\textcolor[rgb]{0,0.0,0}}
\newcommand{\jcth}{\textcolor[rgb]{0,0.,0.}}
\newcommand{\ftfour}{\textcolor[rgb]{0.,0.,0}}
\newcommand{\jcfo}{\textcolor[rgb]{0,0,0}}
\newcommand{\ftfive}{\textcolor[rgb]{0,0,0}}
\newcommand{\jcfi}{\textcolor[rgb]{0,0,0}}
\begin{document}


\title{Cross-modal Consensus Network for Weakly Supervised Temporal Action Localization}

\author{Fa-Ting Hong$^{1,3,4,6,*}$, Jia-Chang Feng$^{1,3,4,7,*}$, Dan Xu$^{5}$, Ying Shan$^{3}$, Wei-Shi Zheng$^{1,2,4,\dag}$}

\makeatletter
\def\authornotetext#1{
\if@ACM@anonymous\else
    \g@addto@macro\@authornotes{
    \stepcounter{footnote}\footnotetext{#1}}
\fi}
\makeatother
\authornotetext{Equal Contribution.}
\authornotetext{Corresponding author.}

\affiliation{
 \institution{\textsuperscript{\rm 1}School of Computer Science and Engineering, Sun Yat-sen University, Guangzhou, China}
  \institution{\textsuperscript{\rm 2}Peng Cheng Laboratory, Shenzhen, China}
 \institution{\textsuperscript{\rm 3}Applied Research Center (ARC), Tencent PCG, Shenzhen, China}
 \institution{\textsuperscript{\rm 4}Key Laboratory of Machine Intelligence and Advanced Computing, Ministry of Education, China}
 \institution{\textsuperscript{\rm 5}Department of Computer Science and Engineering, HKUST, HK}
 \institution{\textsuperscript{\rm 6}Pazhou Lab, Guangzhou, China}
  \institution{\textsuperscript{\rm 7}Guangdong Key Laboratory of Information Security Technology, Sun Yat-sen University, Guangzhou , China}
 \country{}
 }
\email{{hongft3, fengjch8}@mail2.sysu.edu.cn,   danxu@cse.ust.hk, yingsshan@tencent.com,  wszheng@ieee.org}

\def\authors{Fa-Ting Hong, Jia-Chang Feng, Dan Xu, Ying Shan, Wei-Shi Zheng}

\renewcommand{\shortauthors}{Fa-Ting Hong and Jia-Chang Feng et al.}



\begin{abstract}

Weakly supervised temporal action localization (WS-TAL) is a challenging task \ftfour{that aims to localize action instances in \jcfo{the given} video with video-level categorical supervision.}
\jcfi{Previous works use the appearance and motion features extracted from pre-trained feature encoder directly, \eg feature concatenation or score-level fusion. 
 }
\fttwo{In this work, we argue that the features extracted from the pre-trained extractors, \eg I3D, which are
\jcfi{trained for trimmed video action classification, but not specific for WS-TAL task, leading to inevitable redundancy \jcfi{and sub-optimization }. 
}
\jcfi{Therefore,} the feature re-calibration is needed for reducing the task-irrelevant information redundancy.}
\jcfi{Here}, we propose a cross-modal 
\ftone{\jcone{consensus} network (CO$_2$-Net) to tackle this problem.} 
\fttwo{In CO$_2$-Net, we mainly introduce \ftthree{two identical} proposed cross-modal consensus modules (CCM) that design a cross-modal attention mechanism to filter out the task-irrelevant information redundancy using the global information from \jcfo{the} main modality and the cross-modal local information \jcfo{from the} auxiliary modality.}
\fttwo{Moreover, we \jcfi{further explore inter-modality consistency, where we treat} the attention weights derived from each CCM as the pseudo \jctwo{targets} of \ftthree{the attention weights} derived from another CCM to maintain the consistency between the predictions derived from two CCMs, forming a mutual learning manner.}
Finally, we conduct extensive experiments on \jcfo{two commonly used temporal action localization datasets,} THUMOS14 and ActivityNet1.2, to verify our method, \jcfi{which we} achieve \jcfo{the} state-of-the-art results. The experimental results show that our proposed \ftfour{cross-modal consensus module} can produce more representative features for temporal action localization. 

\begin{figure}[ht]
    \centering
    \includegraphics[width=\linewidth]{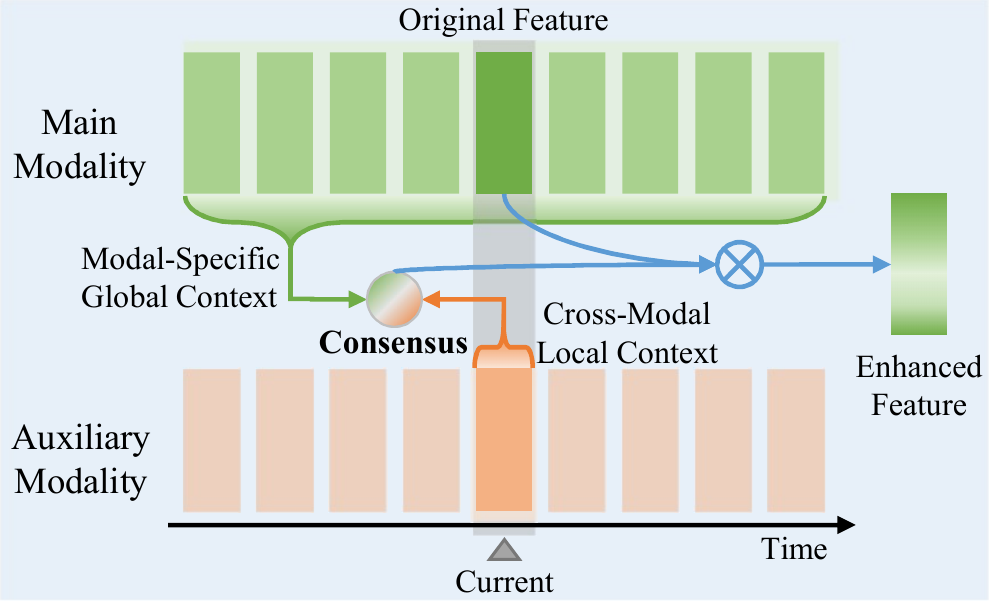}
    \caption{
    \jctwo{Our proposed cross-modal consensus module first encodes modal-specific global context from \jcfo{the} main modality and \jcfo{the} cross-modal \ftfour{local-focused information} from \jcfo{the} current snippet of \jcfo{the} auxiliary modality, which then cooperate to achieve a consensus in modeling channel-wise feature responses and \jcth{enhance the features via information redundancy filtering}}.
    }
    \label{fig:firstpage}
\end{figure}

\end{abstract}




\maketitle

\section{Introduction}\label{sec:intro}
Temporal action localization is a task to localize the start and end timestamps of action instances and recognize their categories. In recent years, many works \cite{zhai2020two, nawhal2021activity, zhao2017temporal,zeng2019graph} put effort into the fully supervised manner and gain great achievements. However, these fully supervised methods require extensive manual frame/snippet level annotations. To address this problem, many weakly supervised temporal action localization (WS-TAL) methods \cite{shou2018autoloc, luo2020weakly,zeng2020hybrid,islam2020weakly,jain2020actionbytes} are proposed to explore an efficient way to detect the action instances in the given videos with only video-level supervision which is more easily obtained by the annotator.

\jcfi{As other weakly supervised video understanding tasks likes video anomaly detection \cite{sultani2018real,feng2021mist} and video highlight detection \cite{hong2020mini}}, most existing WS-TAL methods develop their framework based on the \jcfo{multiple-instance learning} (MIL) manner \cite{lee2021Weakly,islam2020weakly, zeng2020hybrid, lee2020background, liu2019completeness}. These methods firstly predict the categorical probabilities for each snippet and then aggregate them as the video-level prediction. Finally, they perform the optimization procedure using the given video-level labels. \jcfi{Among them, some works \cite{nguyen2018weakly,lee2020background,zeng2020hybrid,liu2019completeness}} introduce an attention module to improve the ability to recognize the foreground by \jcfi{suppressing} the background parts.  \jcth{For action completeness modeling}, Islam \et \cite{islam2021hybrid} utilize an attention module to drop the most discriminative parts of the video but focus on the less discriminative ones. With regards to feature learning, most of WS-TAL methods \cite{islam2020weakly, paul2018w} mainly apply a contrastive learning loss on their intermediate features. Lee \et \cite{lee2021Weakly} \jcfi{proposed to distinguish foreground from background via the inconsistency of their feature magnitudes.} 
\fttwo{The aforementioned methods use the original extracted features that contain the task-irrelevant information redundancy \cite{wang2015learning,wang2019pruning,feng2021mist,lei2021less} to produce predictions directly for each snippet. \jcfi{However, as the features extracted from trained for another task, \ie trimmed video action classification, which introduces redundancy inevitably, } their performances are \jcfo{restricted} to the quality of extracted features \jcfi{and only acquire sub-optimization} \cite{feng2021mist,lei2021less}. \jcfi{Intuitively, performing feature re-calibration for task-specific features is a way to tackle this problem. Instead of finetuning the feature extractor \cite{feng2021mist,alwassel2020tsp,xu2020boundary} with high time and computation cost, we explore to re-calibrate the features 
in a more efficient manner.} 
In this work, our intuition is simple: the RGB and FLOW features contain modal-specific information (\ie appearance and motion information) from different perspectives of the given data. Therefore, we can filter out the redundancy contained in a certain modality with the help of global context information from itself and the \jcth{local context} information from different perspectives of different modalities (Figure~\ref{fig:firstpage}).}


\jcfi{As discussed above, the inconsistency between pre-trained task with the target one leads to inevitable task-irrelevant information in the extracted features denoted as \textit{redundancy}, which restricts the optimization, especially under weak supervision. Previous works pay less attention to this problem but use the features directly.} 
Here, we aim to re-calibrate the features in the very beginning by leveraging \jcth{\ftfour{information} from} two different modalities (\ie RGB and FLOW features).
\fttwo{In this work, we develop a \textbf{C}r\textbf{O}ss-modal c\textbf{O}nsensus \textbf{NET}work (CO$_2$-Net) to re-calibrate the representations of each modality for each snippet in the video. CO$_2$-Net contains two identical cross-modal consensus modules (CCM). Specifically, two types of modal features are fed into \ftfour{both CCMs}, \ftfour{one of them} acts as \jcfo{the} main modality and the \ftfour{other one} serves as \jcfo{the} auxiliary modality. In CCM, we obtain the modality-specific global context information from \jcfo{the} main modality and the cross-modal local-focused descriptor from \jcfo{the} auxiliary modality. Then we aggregate them to produce a channel-wise descriptor that can be used to filter out the task-irrelevant information redundancy. Intuitively, \ftthree{with} the global information of \jcfo{the} main modality, CCM can use the information from different perspectives of the auxiliary modality to determine whether a certain part of the main modality is task-irrelevant information redundancy. Thus we obtain the RGB-enhanced features and FLOW-enhanced features from \ftthree{two} CCMs after filtering the \ftthree{redundancy in original RGB features and FLOW features}, respectively. Then we utilize these two enhanced features to estimate the modality-specific attention weights, respectively, and apply mutual learning loss on these two estimated attention weights for mutual promotion. In addition, we also apply the 
top-k multiple-instance learning loss \cite{paul2018w, islam2020weakly, lee2021Weakly} that \jcfo{is} widely used to learn the temporal class activation map (T-CAM) for each video.}
Finally, we conduct extensive experiments on two public temporal action localization benchmarks, \ie THUMOS14 dataset \cite{THUMOS14} and Activity1.2 dataset \cite{caba2015activitynet}. In our experiments, we investigate and discuss the effect of our proposed cross-modal consensus module with other feature fusion manners (\eg additive  and concatenate function). The experimental results show that our CO$_2$-Net achieves \jcfo{the} state-of-the-art performance on two public datasets, \jcfo{which} verify its efficacy for temporal action localization. To summarize, our contribution is three-fold:
\begin{itemize}
  
    \item\jcfi{As far as we know, it is the first work to investigate multimodal feature re-calibration and modal-wise consistency via mutual learning for temporal action localization.}
    \item \ftthree{We propose a framework, \ie CO$_2$-Net, for temporal action localization to explore a novel cross-modal attention mechanism to re-calibrate the representation for each modality.}

    \item We conduct extensive experiments on two public benchmarks, where our proposed method achieves \jcfo{the} state-of-the-art results.
    
\end{itemize}

\begin{figure*}[t]
    \centering
    \includegraphics[width=0.9\linewidth]{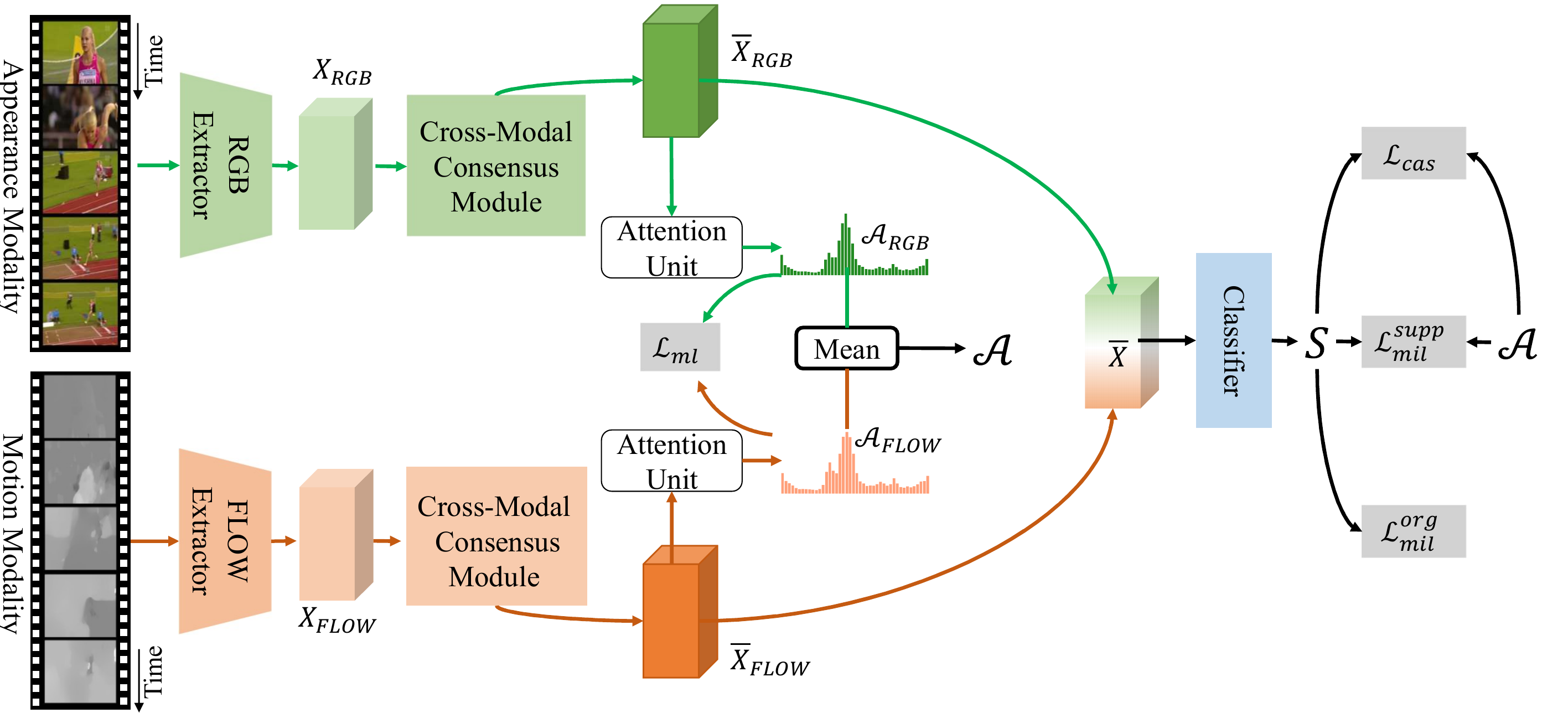}
    \vspace{-1pt}
    \caption{An overview of the proposed cross-modal consensus network (CO$_2$-Net) with two identical CCMs. CCM would \jctwo{filter out the task-irrelevant redundancy of \jcfo{the} main modality and generate the enhanced features for \jcfo{the} main modality by the consensus of both global-context information of itself and the local information from \jcfo{the} auxiliary modality. }
    \ftone{The enhanced features would be fed into the attention unit to estimate the modality-specific attention weights.} On the one hand, we aggregate two attention weights to generate final attention weights $\mathcal{A}$, while these two modality-specific attention weights are optimized by mutual learning loss $\mathcal{L}_{ml}$ for mutual promotion. On the other hand, we first fuse the two  enhanced features as fused features and then feed  \jcfo{them} into a classifier to predict a temporal class activation map (T-CAM). Finally, we apply \jcfo{the} top-k multiple-instance learning loss (\ie $\mathcal{L}_{mil}^{org}$ and $\mathcal{L}_{mil}^{supp}$) and co-activity similarity loss (\ie $\mathcal{L}_{cas}$) to optimize the whole framework.}
    \label{fig:framework}
\end{figure*}
\vspace{-0.3cm}
\section{Related Works}
\noindent\textbf{Weakly Supervised Temporal Action Localization.} Weakly supervised temporal action localization provides an efficient way to detect the action instances without overload annotations. Many works mainly tackle this problem using the multiple-instance learning (MIL) framework \cite{islam2021hybrid, islam2020weakly,lee2020background, lee2021Weakly, liu2021acsnet, luo2020weakly, nguyen2018weakly}. Several works \cite{paul2018w, islam2020weakly} mainly aggregate \ftfour{snippet-level} class scores to produce video-level predictions and learn from video-level action labels. In this formulation, background frames are forced to be mis-classified as action classes to predict video-level labels accurately. To address such a problem, many works \cite{lee2020background,islam2021hybrid} apply an attention module in their framework to suppress the activation of background frames to improve localization performance. Lee \et \cite{lee2020background} introduces an auxiliary class for background and proposes a two-branch weight-sharing architecture with an asymmetrical training strategy.  Besides, MIL-based methods only focus on optimizing the most discriminative snippets in the video \cite{choe2019attention,feng2021mist}. \jcth{For action completeness modeling}, some works \cite{islam2021hybrid, min2020adversarial} adopt the complementary learning scheme that drops the most discriminative parts of the video but \ftfour{focuses} on the complementary parts. Also, several works \cite{pardo2021refineloc, zhai2020two} attempt to optimize their framework under a self-training regime. Zhai \et \cite{zhai2020two} treats the outputs in the last epoch as pseudo labels and refines the network using these pseudo labels.

Different from aforementioned methods, 
\fttwo{this work is the first one that considers filtering out the task-irrelevant information redundancy from each modality with the help of \jctwo{the consensus} of different modalities. Our method aims to \jcfo{re-calibrate} the representation, \ftfour{so that each modality has less information redundancy, which can produce more accurate predictions.}}

\noindent\textbf{Modalities Fusion.} Recently, deep neural networks have been  exploited in multi-modal clustering issue due to powerful feature transformation ability. Many computer vision models \cite{hong2020mini,xu2015learning, deng2018triplet,xu2018PAD-Net, munro2020multi, rao2020a,jing2020cross, xu2017learning} adopt multiple modalities in their framework to obtain performance gains. Different modalities can help to complement each other \ftfour{in} a proper way. In the early stage, Ngiam \et \cite{ngiam2011multimodal} take deep auto-encoder network architecture to learn the common representations of multi-modal data and \ftfour{achieves significant} performance in speech and vision tasks. Several works \cite{hong2020mini, afouras2020self} combine the visual modality and audio modality to tackle a specific task. In general, the video and audio contain different modal information but can enhance each other because visual and audio events tend to occur together. Hong \et \cite{hong2020mini} utilize audio modality in a multiple-head structure to assist vision modality in localizing the video highlights. 

\fttwo{In this work, \jcfi{instead of feature extractor finetuning, }we attempt to filter out the task-irrelevant information redundancy from \ftfour{the specific} modality \jcfi{via a novel re-calibration way,} \jctwo{\jcfi{which we} make a consensus between the global context from itself and }the \jcth{local context} information from \ftfour{another modality}, while the aforementioned works treat the multiple modalities information equally.}

\vspace{-0.27cm}
\section{Method}
\jcone{Video is a typical \jcfo{type of} multimedia that can be translated into multiple modalities that represent the information from different perspectives. } 
\fttwo{In this work, we propose a cross-modal consensus network (CO$_2$-Net) to re-calibrate the representations of each modality using the information from different perspectives of different modalities.}
\vspace{-0.1cm}
\subsection{Problem Formulation} \label{sec:formulation}
We first formulate the WS-TAL problem as follows: suppose $\mathcal{V}$ denotes a batch of data with |$\mathcal{V}$| videos and corresponding video-level categorical labels \fttwo{are} $\mathcal{Y}$, where $\mathcal{Y}=\{Y^{(1)},...,Y^{(|\mathcal{V}|)}\}$ and  $Y^{(i)}=\{y^{(i)}_1, ..., y_C^{(i)}\}=\{0,1\}^C$ for $i$-th video, where $C$ means the number of category. The goal of WS-TAL is to learn a function that  simultaneously detects and classifies all action instances temporally with precise timestamps as ($t_s,t_e,c,\gamma$) for each video, where $t_s$,$t_e$,$c$,$\gamma$ denote the start time, the end time, the predicted category and the confidence score for corresponding action proposal, respectively.

\vspace{-0.1cm}
\subsection{Pipeline}
\noindent\textbf{Feature Extraction.}
 Following recent WS-TAL methods \cite{paul2018w, islam2021hybrid}, we construct CO$_2$-Net upon snippet-level feature sequences extracted from non-overlapping video volumes, where each volume contains 16 frames. The features for appearance modality (RGB) and motion modality (optical flows) are both extracted from pretrained extractors, \ie I3D \cite{carreira2017quo}. The features for appearance and motion modality are 1024-dimension for each snippet. For $i$-th video with $T$ snippets, we use matrix tensors $X_{RGB} \in \mathbb{R}^{T\times D}$ and $X_{FLOW} \in \mathbb{R}^{T\times D}$ to represent the RGB and FLOW features of the whole video, respectively, where D means the dimension of \jcth{\ftfour{the} feature vector}.

\noindent\textbf{Structure Overview.}
Figure \ref{fig:framework} shows the whole pipeline of our proposed CO$_2$-Net.
Both RGB and FLOW features are fed in two \jcfo{identical} cross modal consensus modules. 
\ftthree{In each CCM, \jcfo{we select one of the two modalities }
as \jcfo{the} main modality \jcfo{that} will be enhanced by removing the task-irrelevant information redundancy \jcfo{with the help of} the global context of itself and cross-modal local-focused information from another (auxiliary) modality.}
Thus we can obtain the more task-specific representation for each modality.
Then, the enhanced representation is utilized to produce attention weights \jcfo{that indicate} the probabilities of \ftfour{each snippet being foreground through an attention unit that consists of two convolution layers.} We aggregate two attention weights generated by enhanced features from two CCMs respectively to produce final attention weights that can be used in the testing stage. And we also fuse the two enhanced features and feed them into a classifier to predict the categorical probabilities for each snippet. 


\vspace{-0.1cm}
\subsection{Cross-modal Consensus Module} \label{sec:CCM}

\begin{figure}[t]
    \centering
    \includegraphics[width=\linewidth]{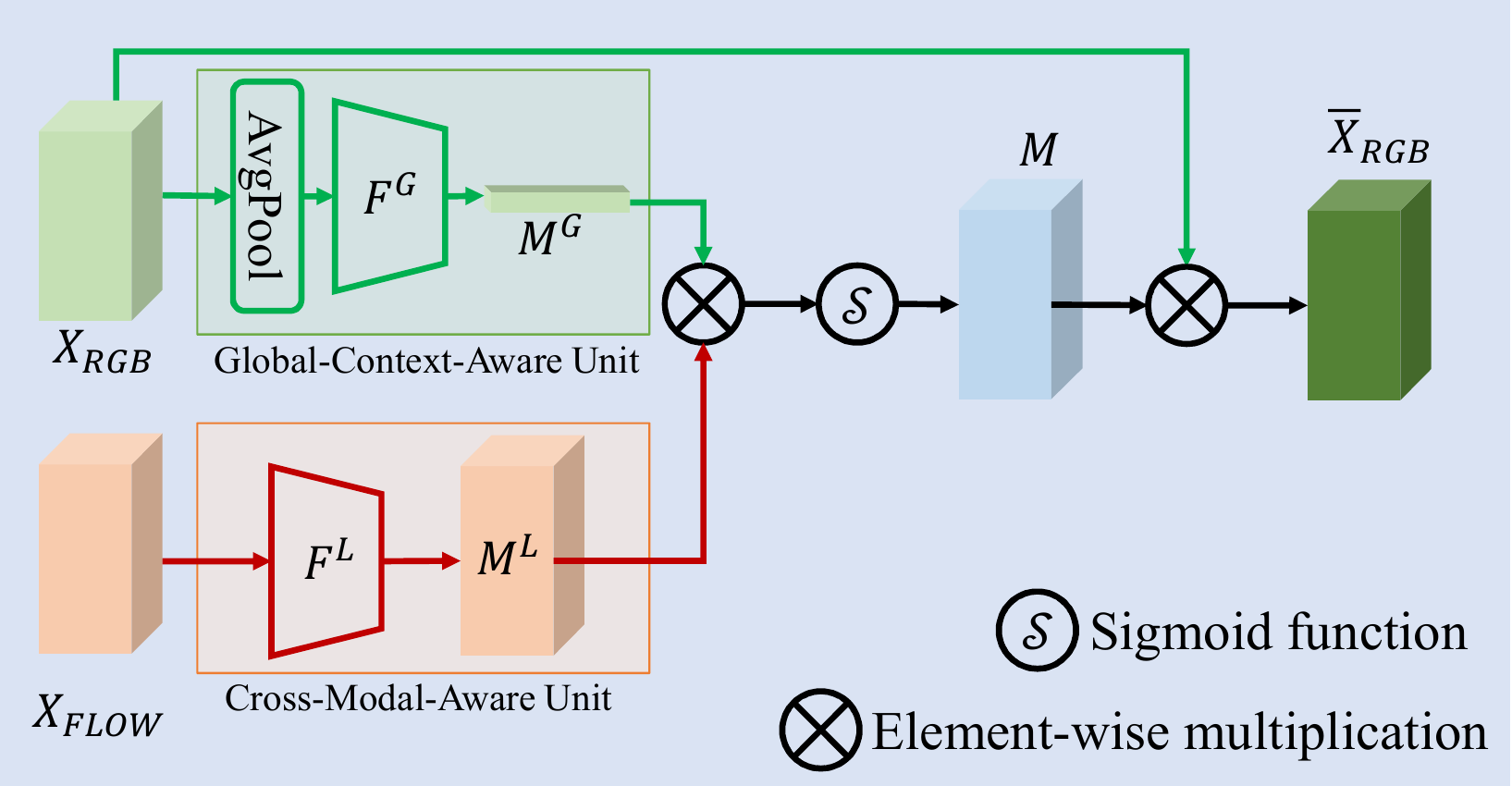}
    \caption{An overview of the proposed cross-modal consensus module. \jcone{The module contains a Global-Context-Aware unit and Cross-Modal-Aware unit to distinguish the information redundancy and re-calibrate the features. }
    \jcone{In this module, the main modality ${X}_{RGB}$ cooperates with the auxiliary modality ${X}_{FLOW}$ to generate channel-wise descriptors $M$ to govern the excitation of each channel to filter out the information redundancy. Thus the main modality features are then enhanced by channel-wise attention mechanism as $\overline{X}_{RGB}$. The workflow is same when the roles of these modalities are exchanged.   
    }
    }
    \label{fig:CCM}
    \vspace{-0.3cm}
\end{figure}

In this work, we employ a cross-modal consensus module to \jcone{filter out the task-irrelevant information redundancy for each modality before the process of downstream learning task. \ftfour{The proposed} cross-modal consensus module is constructed by a global-context-aware unit and a cross-modal-aware unit to distinguish the information redundancy and filter out them via a channel-wise suppression on the features. As shown in Figure \ref{fig:CCM}, we treat the appearance modality (RGB features) as \jcfo{the} main modality and the motion modality (FLOW features) as the auxiliary }to feed in our proposed cross-modal consensus module, while the same workflow is performed when the roles of the two modality are exchanged. \jctwo{For the convenience of expression, we take RGB features as the main modality features as an example in the rest of the article.}


\fttwo{As the features are extracted from a encoder that pretrained on some large datasets not related to WS-TAL task, thus the features may \ftfour{contain} some task-irrelevant misleading redundancy that restricts the localization performance.}
Given \jcfo{the} main modality and \jcfo{the} auxiliary modality, instead of \jcone{directly concatenating them, we aim to design a mechanism to filter out the task-irrelevant information redundancy in \jcfo{the} main modality. }
Motivated by the self-attention mechanism \cite{vaswani2017attention} \jcone{and squeeze-and-excitation block \cite{hu2018squeeze}, we develop a similar manner, named cross-modal attention mechanism, to distinguish the information redundancy and filter out them.} 

\jcone{In the global-context-aware unit, we first squeeze modality-specific global context information into a video-level feature $X_g \in \mathbb{R}^D$, which is aggregated from the main modality $X_{RGB}$, using an average pooling operator $\psi(\cdot)$ \fttwo{on temporal dimension}. Then, we adopt a convolution layer $F^G$ to fully capture channel-wise dependencies and produce modality-specific global-aware descriptor $M^G$. The process is formulated below: }
\begin{equation} \label{eq:gca}
    \begin{aligned}
    X_g &= \psi(X_{RGB}),\\
    M^G &= F^G(X_g).
    \end{aligned}
\end{equation}

\fttwo{As multiple modalities provide information from different perspectives, we can leverage the information from \jcfo{the} auxiliary modality to detect the task-irrelevant information redundancy in \jcfo{the} main modality.
 Thus, in the cross-modal-aware unit, we aim to capture the cross-modal local-specific information from the auxiliary modality features $X_{FLOW}$. 
 \jcfo{Here,} we introduce a convolution layer $F^L$ that 
 embed the features of \jcfo{the} auxiliary modality 
 to produce a cross-modal local-focused descriptor $M^L$ as follows: }
\begin{align}
   M^L &=F^L(X_{FLOW}). 
\end{align}

\jcone{Here, we obtain channel-wise descriptor $M$ for feature re-calibration by multiplying modality-specific global-aware descriptor $M^G$ with cross-modal local-focused descriptor $M^L$. Finally, the task-irrelevant information redundancy is filtered out via a \ftfour{cross-modal attention mechanism} as follows: }
\begin{equation}\label{eq:CCM}
    \begin{aligned}
        M &= M^G\otimes M^L, \\
        \overline{X}_{RGB} &= \sigma(M)\otimes X_{RGB},
    \end{aligned}
\end{equation}
where $\sigma(\cdot)$ is a Sigmoid function\jcfo{, while the ``$\otimes$'' means element-wise multiplication operator}. Remarkably, $M^G$ and $M^L$ can be treated as ``Query'' and ``Key'' in the self-attention module \cite{vaswani2017attention}. Instead of using a softmax operator, we apply a Sigmoid function to produce channel-wise re-calibration weights to enhance the original main modality features $X_{RGB}$.


\vspace{-0.1cm}
\subsection{Dual Modal-specific Attention Units}

After obtaining the enhanced features, we attempt to produce modality-specific temporal attention weights that indicate the snippet-level probabilities of being foreground. Here, following previous works \cite{lee2020background, islam2021hybrid}, we feed the enhanced features into the attention unit $F^A_{RGB}$ for modality-specific attention weights:

\begin{equation}
    \mathcal{A}_{RGB} = F^A_{RGB}(\overline{X}_{RGB}),
\end{equation}
where $F^A_{RGB}(\cdot)$ is the attention unit for RGB with three convolution layers, \jctwo{which is same with the attention unit for FLOW $F^A_{FLOW}$}. 

As we have two CCMs in the proposed CO$_2$-Net, we obtain the RBG-enhanced features $\overline{X}_{RGB}$ and modality-specific attention weights $\mathcal{A}_{RGB}$ from one CCM that treats the appearance modality as the main modality and motion modality as \jcfo{the} auxiliary modality, while we also gain the FLOW-enhanced features $\overline{X}_{FLOW}$ and modality-specific attention weights $\mathcal{A}_{FLOW}$ from another CCM, \ftfour{in which} the roles of two modalities are opposite to the former CCM.

After obtaining the enhanced features (\ie $\overline{X}_{RGB}$ and $\overline{X}_{FLOW}$) and modality-specific attention weights (\ie $\mathcal{A}_{RGB}$ and $\mathcal{A}_{FLOW}$). We first fuse two attention weights:
\begin{equation}
    \mathcal{A} = \frac{\mathcal{A}_{RGB}+\mathcal{A}_{FLOW}}{2}.
\end{equation}

We think that the two modality-specific attention weights produced by two enhanced features respectively have different emphasis on the video, while the fused attention weights $\mathcal{A}$ can better represent the probability of \ftthree{snippet} being foreground because it made a trade-off between the two modality-specific attention weights. 
Finally, We concatenate two types of enhanced features, \ie $\overline{X}_{RGB}$ and $\overline{X}_{FLOW}$, to form $\overline{X}$ and feed it into a classifier that contains three convolution layers to produce the temporal class activation map (T-CAM) $\mathcal{S} \in \mathbb{R}^{T \times (C+1)}$ for the given video, where the $(C + 1)$-th class is the background class. 

\begin{figure}
    \centering
    \includegraphics[width=\linewidth]{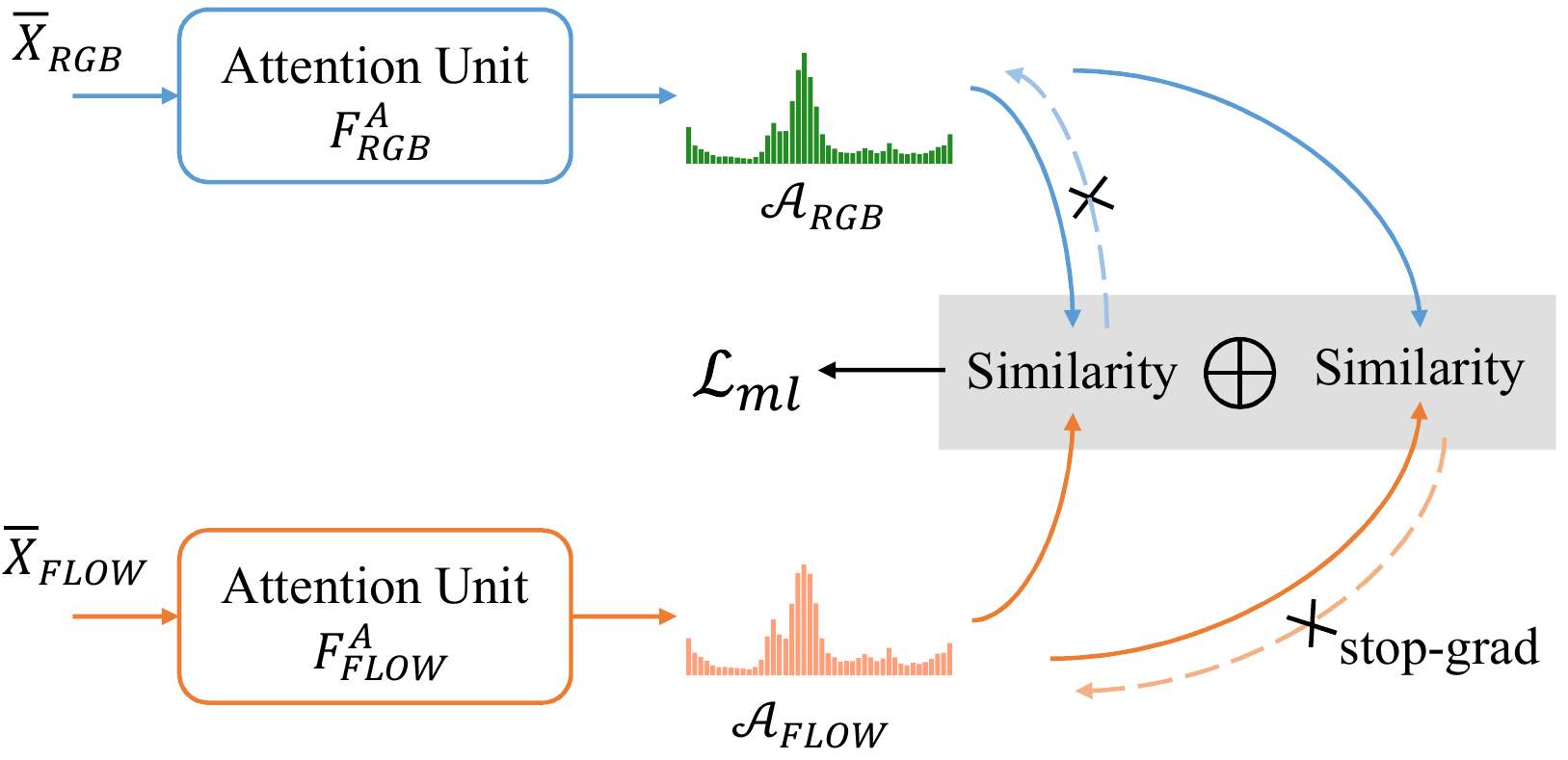}
    \vspace{-0.2cm}
    \caption{Illustration of the workflow of the mutual learning process. The two temporal attention weights generated from dual model-specific attention units are learning from each other by treating the other as pseudo labels and stopping the gradients backwards. }
    \label{fig:mutual_learning}
    \vspace{-0.5cm}
\end{figure}

\subsection{Optimizing Process}
\noindent\textbf{Constraints on Attention Weights.}
Here, we have obtained two modality-specific attention weights (\ie $\mathcal{A}_{RGB}$ and $\mathcal{A}_{FLOW}$) and a fused attention weights $\mathcal{A}$. Then we first apply mutual learning scheme on two modality-specific attention weights:
\begin{equation}\label{eq:ml}
    \mathcal{L}_{ml}=\alpha \delta(\mathcal{A}_{RGB},\phi(\mathcal{A}_{FLOW}))+(1-\alpha) \delta(\mathcal{A}_{FLOW},\phi(\mathcal{A}_{RGB})),
\end{equation}
where $\phi(\cdot)$ represents a function that truncates the gradient of input, \jcone{while $\delta(\cdot)$ means a similarity metric function} \ftfour{and $\alpha$ is a hyperparameter}. In Eq. \ref{eq:ml}, we treat $\mathcal{A}_{RGB}$ and $\mathcal{A}_{FLOW}$ as pseudo-labels of each other \ftone{(as shown in Figure \ref{fig:mutual_learning})}, so that they can learn from each other and align the attention weights. \jcone{Here, we adopt mean square error (MSE) as function $\delta(\cdot)$ in Eq. \ref{eq:ml}}. \ftone{\jctwo{Besides} MSE, we also discuss others similarity metric functions (\ie Jensen-Shannon (JS) divergence, Kullback-Leibler (KL) divergence and mean absolute error (MAE) ) that \jcfo{is applied} in Eq. \ref{eq:ml} in Section \ref{sec:abla}.}
In addition, we can find that the distribution of attention weights should be opposite to the probability distribution of the background class in $\mathcal{S}$:
\begin{equation}
    \mathcal{L}_{oppo} = \frac{1}{3}( |\mathcal{A}_{RGB}+s_{c+1}-1|+|\mathcal{A}_{FLOW}+s_{c+1}-1|+|\mathcal{A}+s_{c+1}-1|),
\end{equation}
where $|\cdot|$ is a absolute value function, and $s_{c+1}$ is the last column in the \ftfour{T-CAM} $\mathcal{S}$ that represents the probabilities of each snippet being background. And we also utilize a normalization loss $\mathcal{L}_{norm}$ to make the attention weights more polarized:
\begin{equation}
    \mathcal{L}_{norm} = \frac{1}{3}( ||\mathcal{A}_{RGB}||_1+||\mathcal{A}_{FLOW}||_1+||\mathcal{A}||_1),
\end{equation}
where $||\cdot||_1$ is a L1-norm function.

\noindent\textbf{Constraints on T-CAMs and Features.}
In order to better recognize the background activity, we apply the attention weights $\mathcal{A}$ to suppress the background snippets in T-CAM $\mathcal{S}$ and obtain suppressed T-CAM $\overline{\mathcal{S}}$:
\begin{equation}
    \overline{\mathcal{S}} = \mathcal{A} \otimes \mathcal{S}.
\end{equation}

In this work, we apply \jcfo{the} widely used top-k multiple-instance learning loss \cite{paul2018w} on T-CAM $\mathcal{S}$ and $\overline{\mathcal{S}}$, denoted as $\mathcal{L}_{mil} = \mathcal{L}_{mil}^{org} + \mathcal{L}_{mil}^{supp}$. Also, we apply the co-activity similarity loss $\mathcal{L}_{cas}$\footnote{Both the top-k multiple-instance learning loss and co-activity similarity are widely used in current WS-TAL methods. They are not the main contributions in this work, so that we do not detail them in our paper. More details of them can refer to \cite{paul2018w}.} \cite{paul2018w} on fused features $\overline{X}$ and suppressed T-CAM $\overline{\mathcal{S}}$ to learn better representations and T-CAM. Because we utilize the suppressed T-CAM in \jcfo{the} testing stage in Section \ref{sec:local}, we only apply $\mathcal{L}_{cas}$ on suppressed T-CAM.

\noindent\textbf{Final Objective Function. }
Finally, we aggregate all aforementioned objective functions to form the final objective function for whole framework optimization:
\begin{equation} \label{eq:final}
    \mathcal{L} = \mathcal{L}_{mil}+\mathcal{L}_{cas}+\mathcal{L}_{ml}+\lambda_1\mathcal{L}_{oppo} + \lambda_2\mathcal{L}_{norm},
\end{equation}
here, the \jcth{$\lambda_1$ and $\lambda_2$} are hyperparameters. Our framework can learn more robust representation to produce more accurate T-CAM by optimizing that final objective function.

\vspace{-0.1cm}
\subsection{Temporal Action Localization} \label{sec:local}
At the testing stage, we follow the process of \cite{islam2021hybrid}. Firstly, we calculate the video-level categorical probabilities that indicate the possibility of each action class happened in the given video. Then we set a threshold $\tau$ to determine the action classes that would be localized in the video. For the selected action class, we threshold the attention weights $\mathcal{A}$ to drop the background snippets and obtain the class-agnostic action proposals by selecting the continuous components of the remaining snippets. As we said in Section \ref{sec:formulation}, a candidate action proposal is a four-tuple: ($t_s,t_e,c,\gamma$). After obtaining the action proposals, we utilize the suppressed T-CAM $\overline{\mathcal{S}}$ to calculate the class-specific score $\gamma$ for each proposal using Outer-Inter Score \cite{shou2018autoloc}. Moreover, we use multiple thresholds to threshold the attention weights to enrich the proposal set with proposals in 
different levels of scale. 
\jcth{ Further, we remove the overlapping proposals \ftfour{using} soft non-maximum suppression.}

\vspace{-0.1cm}
\section{Experiments}
In this section, we conduct extensive experiments on two public temporal action localization benchmarks, \ie THUMOS14 \cite{THUMOS14} and ActivityNet1.2 dataset \cite{caba2015activitynet}, to investigate the effectiveness of our proposed framework. In addition, we conduct ablation studies to discuss each component in CO$_2$-Net and visualize some results.
\begin{table*}[t] 
	\centering
	\scalebox{0.9}
	{
	\begin{tabular}{c|c||ccccccccc|ccc}
	    \toprule
	    \multirow{2}{*}{Supervision}  & \multirow{2}{*}{Method}    & \multicolumn{9}{c}{mAP@IoU (\%)} & \multicolumn{3}{|c}{AVG mAP (\%)}\\
	    \cline{3-14}
	    & & 0.1 & 0.2 & 0.3 & 0.4 & 0.5 & 0.6 & 0.7 & 0.8 & 0.9 &0.1:0.5 &0.1:0.7 & 0.1:0.9\\ 
	    \midrule
	    \midrule
	    \multirow{5}{*}{Fully} 
	     & S-CNN \cite{shou2016temporal} (2016) & 47.7 & 43.5 & 36.3 & 28.7 & 19.0 & 10.3 & 5.3 & - & -&35.0& 24.3 &- \\
	     & SSN\cite{zhao2017temporal} (2017)  & 60.3 & 56.2 & 50.6 & 40.8 & 29.1	& - & - & - & - &47.4& -&-\\
	     & BSN \cite{lin2018bsn} (2018) & - & - & 53.5 & 45.0 & 36.9 & 28.4 & 20.0 &  - & - & -&- &-\\
	     & TAL-Net \cite{chao2018rethinking} (2018)  & 59.8 & 57.1 & 53.2 & 48.5 & 42.8 & \textbf{33.8} & \textbf{20.8}& - & - & 52.3 & 45.1 &-\\
	    & P-GCN\cite{zeng2019graph} (2019) & 69.5 & \textbf{67.5} & \textbf{63.6} & \textbf{57.8} & \textbf{49.1} & - & - &  -& -&\textbf{61.5} & -&- \\
        \midrule
        \multirow{5}{*}{Weakly\dag} 
        & CMCS\cite{liu2019completeness} (2019) & 57.4 & 50.8 & 41.2 & 32.1 & 23.1 & 15.0 & 7.0 & - & -&40.9 &32.4 &-\\
        & STAR \cite{xu2019segregated} (2019) & 68.8	& 60.0 & 48.7 & 34.7 & 23.0	& - & - & - & - &47.4& -&-\\
        & 3C-Net \cite{narayan20193c} (2019)  & 59.1 & 53.5 & 44.2 & 34.1 & 26.6 & - & 8.1 &  -& - &43.5& - &-\\
        & PreTrimNet \cite{zhang2020multi} (2020) & 57.5	& 50.7 & 41.4 & 32.1 & 23.1	& 14.2 & 7.7 & - & -& 41.0 & 23.7 &-\\
        & SF-Net \cite{ma2020sf} (2020) & 71.0 & 63.4 & 53.2 & 40.7 & 29.3 & 18.4 & 9.6 & - & -& 51.5 & 40.8 &-\\
        \midrule
        
        \multirow{9}{*}{Weakly} 
        & BaS-Net \cite{lee2020background} (2020)& 58.2 & 52.3 & 44.6 & 36.0 & 27.0 & 18.6 & 10.4& 3.3 & 0.4 & 43.6 &35.3 &27.9\\
        & Gong \et \cite{gong2020learning} (2020)& - & - & 46.9 & 38.9 & 30.1	& 19.8 & 10.4 &  -& - & -&- &-\\
        & DML \cite{islam2020weakly} (2020) & 62.3 & - & 46.8 & - & 29.6 & - & 9.7 &  - & - & -&- &-\\
        & A2CL-PT \cite{min2020adversarial} (2020) & 61.2 & 56.1	& 48.1 & 39.0 & 30.1 & 19.2 & 10.6 & 4.8 & 1.0 &46.9& 37.8 &30.0\\
        & TSCN \cite{zhai2020two} (2020)& 63.4 & 57.6 & 47.8 & 37.7 & 28.7 & 19.4 & 10.2 & 3.9 & 0.7 &  47.0&37.8 &29.9\\
        & ACSNet \cite{liu2021acsnet} (2021) & - & - & 51.4 & 42.7 & 32.4	& 22.0 & 11.7 &  - & - & -&-&-\\

        & HAM-Net \cite{islam2021hybrid} (2021) & 65.9	& 59.6 & 52.2 & 43.1 & 32.6	& 21.9 & 12.5& 4.4* & 0.7* &  50.7& 39.8 &32.5\\
        & UM \cite{lee2021Weakly} (2021) & 67.5	& 61.2 & 52.3 & 43.4 & 33.7	& 22.9 & 12.1 & 3.9* & 0.4* &  51.6& 41.9& 33.0 \\
        & \textbf{CO$_2$-Net} & \textbf{70.1} & \textbf{63.6} & \textbf{54.5} & \textbf{45.7} & \textbf{38.3} & \textbf{26.4} & \textbf{13.4} & \textbf{6.9} & \textbf{2.0} & \textbf{54.4} & \textbf{44.6}  & \textbf{35.7} \\
        
        \bottomrule
	\end{tabular}
		
	}
    \centering\caption{Comparisons of CO$_2$-Net with other methods on the THUMOS14 dataset. AVG is the average mAP under multiple thresholds, namely, 0.1:0.5:0.1, 0.1:0.7:0.1 and 0.1:0.9:0.1; while $\dag$ means additional information is adopted in this method, such as action frequency or human pose. * indicates \ftfour{that the results} are obtained by contacting the corresponding authors via email.}
    \label{tab:comp_thumos}
 	\vspace{-0.2cm}
\end{table*}%

\vspace{-0.1cm}
\subsection{Datasets and Metrics}
We evaluate our proposed approach on two public benchmark datasets, \ie THUMOS14 dataset \cite{THUMOS14} and ActivityNet1.2 dataset \cite{caba2015activitynet}, for temporal action localization.

\noindent\textbf{THUMOS14.} There are 200 validation videos and 213 test videos of 20 action classes in THUMOS14 dataset. These videos have diverse length and those actions frequently occur in the videos. Following the previous works \cite{islam2021hybrid,paul2018w}, we use 200 validation videos to train our framework and 213 test videos for testing.

\noindent\textbf{ActivityNet1.2.} ActivityNet1.2 dataset is a large temporal action localization dataset with coarser annotations. It is composed of 4,819 training videos, 2,383 validation videos and 2,489 test videos of 100 action classes. We cannot obtain the ground-truth annotations for the test video, because they are withheld for the challenge. Therefore, we utilize validation videos for testing \cite{islam2021hybrid,islam2020weakly}.

\noindent\textbf{Evaluation Metrics.} In this work, we evaluate our method with mean average precision (mAP) under several different intersections of union (IoU) thresholds, which are the standard evaluation metrics for temporal action localization \cite{paul2018w}. Moreover, we utilize the official released evaluation code\footnote{http://github.com/activitynet/ActivityNet} to measure our results.
\vspace{-0.1cm}
\subsection{Implementation Details}
In this work, we implement our method in PyTorch \cite{paszke2019pytorch}. In the very beginning, we apply I3D networks \cite{carreira2017quo} pretrained on Kinetics-400 \cite{kay2017kinetics} to extract both RGB and FLOW features for each video, following previous work \cite{islam2020weakly,paul2018w}. We sample continuous non-overlapping 16 frames from video as a snippet, where the features for each modal of each snippet are 1024-dimension. In the training stage, we randomly sample 500 snippets for THUMOS14 dataset and 60 snippets for ActivityNet1.2 dataset, while all snippets are taken during testing. For fair comparisons, we do not finetune the feature extractor, \ie I3D.
\jcone{The attention unit is constructed with 3 convolution layers, whose  output dimensions are 512, 512 and 1 while the kernel sizes are 3, 3 and 1. The classification module contains 3 temporal convolution layers. Between each convolution layer, we use Dropout regularization with possibility as 0.7. }

For each hyperparameters, we set $\lambda_1 =\lambda_2 = 0.8$ \jcth{for the last two terms of regularization in the final objective function}, and $\alpha=0.5$ \ftthree{to obtain the best performance }for both two datasets. In the training process, we sample 10 videos in a batch, \jcfo{in which there are 3 pairs of videos \ftfour{and each pair contains} the same categorical tags }
for co-activity similarity loss $\mathcal{L}_{cas}$. We deploy Adam optimizer \cite{kingma2014adam} for optimizing, \ftfour{in which} the learning rate is 5e-5 and weight decay rate is 0.001 for THUMOS14, while 3e-5 and 5e-4 for ActivityNet1.2 dataset. \jcth{All experiments are \ftfour{run} on a single NVIDIA GTX TITAN (Pascal) GPU.}

\vspace{-0.1cm}
\subsection{Comparison With State-of-the-art Methods}
We first compare our proposed CO$_2$-Net with current weakly supervised state-of-the-art methods and several fully supervised methods. We report the results in Table \ref{tab:comp_thumos} and Table \ref{tab:comp_act}. From Table \ref{tab:comp_thumos}, we can find that our method outperforms all weakly supervised methods in all IoU metrics \ftfour{on the THUMOS14 dataset}, while even comparable with fully supervised methods at low IoU region. Compared with those native early fusion methods (\eg HAM-Net \cite{islam2021hybrid} and UM \cite{lee2021Weakly}) and late fusion methods (\eg TSCN \cite{zhai2020two}), our method gains a significant improvement. For example, The results on ``AVG mAP (0.1:07)'' of CO$_2$-Net vs. that of UM is 44.6\% vs. 41.9\%. 
\fttwo{These results show that using the information from different modalities to reduce the task-irrelevant information redundancy can benefit the temporal action localization.}
In addition to this, we also compare our method with several fully supervised, we can find that the results produced by our CO$_2$-Net are even comparable with those fully supervised methods in terms of metrics with low IoU, \ie mAP@IoU0.1 and mAP@IoU0.2. Moreover, our method even outperforms some fully supervised methods, \eg S-CNN \cite{shou2016temporal} and BSN \cite{lin2018bsn}. These results validate the effectiveness of our proposed method.

With regard to the results of ActivityNet1.2 dataset reported in Table \ref{tab:comp_act}, we can find that our method is still better than the current SOTA methods \ftfour{on the whole}. However, We can not obtain the same impressive improvement \ftfour{on} ActivityNet1.2 dataset as it we do in the THUMOS14 \ftfour{dataset}, because ActivityNet1.2 dataset has only 1.5 action instances per video, compared with THUMOS14 dataset which has around 15 action instances per video. Additionally, we find that the annotations of ActivityNet1.2 dataset are coarser than those in THUMOS14 dataset. Taking all these into account, we recognize that the THUMOS14 dataset is more suitable for temporal action localization task than ActivityNet1.2 dataset (\ftfive{as discussed in} \cite{islam2020weakly}). Therefore, we mainly use the former to verify our method in the following.

\begin{table}[t] 
	
	\centering
	{
	     \resizebox{1\columnwidth}{!}{
	     \begin{tabular}{c|c||ccc|c}
		    \toprule
		    \multirow{2}{*}{Supervision}  & \multirow{2}{*}{Method} & \multicolumn{4}{c}{mAP@IoU (\%)} \\
		    \cline{3-6}
		    & & 0.5 & 0.75 & 0.95 & AVG \\
		    \midrule
		    \midrule
		    \multirow{1}{*}{Fully}
		    & SSN\cite{zhao2017temporal} (2017) & 41.3 & 27.0 & 6.1 & 26.6 \\

            \midrule
            
            \multirow{2}{*}{Weakly\dag}
            & 3C-Net \cite{narayan20193c} (2019) & 35.4 & 22.9 & 8.5 & 21.1 \\
		    & CMCS \cite{liu2019completeness} (2019)& 36.8 & 22.0  & 5.6 & 22.4 \\
            \midrule
            
            \multirow{10}{*}{Weakly}
            & BaSNet \cite{lee2020background} (2020) & 38.5 & 24.2 & 5.6 & 24.3 \\
            & ActionBytes \cite{jain2020actionbytes} (2020) & 39.4 & - & - & - \\ 
            & DGAM \cite{shi2020weakly} (2020) & 41.0 & 23.5 & 5.3 & 24.4 \\
            & Gong \et \cite{gong2020learning} (2020) & 40.0 & 25.0 & 4.6 & 24.6 \\
		    & TSCN \cite{zhai2020two} (2020) & 37.6 & 23.7 & 5.7 & 23.6 \\
		    & RefineLoc \cite{pardo2021refineloc} (2021) & 38.7 & 22.6 & 5.5 & 23.2 \\
		    & HAM-Net \cite{islam2021hybrid} (2021) & 41.0 & 24.8 & 5.3 & 25.1 \\
		    & UM \cite{lee2021Weakly} (2021) & 41.2 & 25.6 & 6.0 & 25.9 \\
		    & ACSNet \cite{liu2021acsnet} (2021) & 40.1 & 26.1 & \textbf{6.8} & 26.0 \\
		    & \textbf{CO$_2$-Net} & \textbf{43.3} & \textbf{26.3} & 5.2 & \textbf{26.4} \\
            \bottomrule
		\end{tabular}
		}
		
	}
    \centering\caption{Comparison of our algorithm with other methods on the ActivityNet1.2 dataset. AVG means average mAP from IoU 0.5 to 0.95 with 0.05 increment.}
    \label{tab:comp_act}
    \vspace{-0.4cm}
\end{table}%

\vspace{-0.1cm}
\subsection{Ablation study} \label{sec:abla}
In this work, we propose a cross-modal consensus module to \ftfour{re-calibrate the representations and }produce the \ftfour{enhanced features}, and a mutual learning loss to enable two CCMs can learn from each other. Also, our final objective function consists of several components. Here, we first conduct the ablation studies to investigate the effect of each object functions. Then we discuss different kinds of combination of main and auxiliary modalities in the cross-modal consensus module. Finally, we also illustrate the results of different multi-modal fusion methods as well as SE-attention \cite{hu2018squeeze} \ftfour{that replace the CCM in} CO$_2$-Net to verify the effectiveness of CCM.

\begin{table}[t]
    \begin{tabular}{c|ccccc|c}
        \toprule
        Exp & $\mathcal{L}_{mil}$& $\mathcal{L}_{oppo}$ & $\mathcal{L}_{ml}$ & $\mathcal{L}_{cas}$ & $\mathcal{L}_{norm}$ & Avg mAP (\%) \\ \midrule
        1   &  \checkmark      &          &           &                &          & 38.1    \\
        2   &  \checkmark      &  \checkmark        &           &                &          & 40.0    \\
        3   &  \checkmark      &  \checkmark        &  \checkmark         &                &          & 41.4    \\
        4   &  \checkmark      &  \checkmark        &  \checkmark         &   \checkmark             &          & 42.8    \\ 
        5   &  \checkmark      &    \checkmark      &                     &   \checkmark    &
    \checkmark          &   42.6    \\\hline
        6   &  \checkmark      &  \checkmark        &  \checkmark         &   \checkmark             &    \checkmark      & 44.6     \\ 
        \bottomrule
    \end{tabular}
    \caption{Ablation studies of our algorithm in term of average mAP under multiple IoU thresholds as \{0.1:0.7:0.1\}.}
    \label{tab:ablation_loss}
    \vspace{-0.4cm}
\end{table}

\begin{table}[t]
\begin{tabular}{c|ccccccc|c}
\toprule
\multirow{2}{*}{$\mathcal{L}_{ml}$} & \multicolumn{7}{c|}{mAP@IoU} & \multirow{2}{*}{AVG} \\ \cline{2-8}
           & 0.1   & 0.2   & 0.3  & 0.4  & 0.5  & 0.6 & 0.7 &   \\ 
           \midrule
            MAE        &   69.2     &    63.0   &    53.8  &  45.2    & \textbf{38.6}     &  26.2    & 13.9       &   44.3     \\
            KL         &  67.7     &    62.2   &    53.9  & 44.8     & 37.3     &  25.7   & \textbf{14.7}    &   43.8     \\
            JS         &    69.2   &    63.3   &    \textbf{54.5}  & \textbf{46.0}     &  38.3    &   \textbf{26.5}  &  14.1      &  44.5    \\ 
            MSE        &   \textbf{70.1}   &    \textbf{63.6}   &   \textbf{54.5}   &  45.7  & 38.3     &  26.4    & 13.4            &   \textbf{44.6}     \\
\bottomrule
\end{tabular}
\caption{Ablation studies of different types of mutual learning loss in term of average mAP under multiple IoU thresholds from 0.1 to 0.7 with interval as 0.1.}
    \label{tab:ablation_ml}
    \vspace{-0.4cm}
\end{table}

\noindent\textbf{Effect of each component of final objective function.} Each component in the final objective function (Eq. \ref{eq:final}) performs important role in our framework to help to learn the feature representations and final predictions. To verify the effectiveness of each objective function, we conduct related ablation studies and report results in Table \ref{tab:ablation_loss}. We can find that each objective function makes contributions to the final performance. We treat the ``Exp 1'' as our baseline that only \ftfour{uses} multiple-instance learning loss $\mathcal{L}_{mil}$. It is notable that our baseline is similar to the BaS-Net \cite{lee2020background} but our baseline outperforms \ftfour{the latter} by 2.8\%. 
\fttwo{Because our baseline contains the cross-modal consensus module to filter out the task-irrelevant information redundancy from two modalities and \ftfour{uses} the concatenation of two enhanced features as the representation of each snippet.}
The results \ftfour{in} Table \ref{tab:ablation_loss} show that each component in the final objective function can help to train our proposed CO$_2$-Net. We also \ftfour{evaluate} the effect of the different types of mutual learning loss and report the results in Table \ref{tab:ablation_ml}. The results of all types of mutual learning loss can outperform the current state-of-the-art results shown in Table \ref{tab:comp_thumos}. These results indicate that it is necessary for the two CCMs to learn from each other and MSE is more suitable.

\begin{table}[t]
\begin{tabular}{cc|ccccccc||c}
\toprule
\multirow{2}{*}{Main} & \multirow{2}{*}{Auxiliary} & \multicolumn{4}{c||}{mAP@IoU} & \multirow{2}{*}{AVG} \\ \cline{3-6}
& & 0.1  & 0.3  & 0.5  & \multicolumn{1}{c||}{0.7} & \\
\midrule
  Local     & Local     &  68.1    &  52.6    &  36.3    &  \multicolumn{1}{c||}{13.2}  &  43.0      \\
  Local     & Global    &  70.0    &  54.1    &  37.5    &  \multicolumn{1}{c||}{12.3}  &  44.0       \\
  Global    & Local     &  \textbf{70.1}    &  \textbf{54.5}    &  \textbf{38.3}    &  \multicolumn{1}{c||}{\textbf{13.4}}  &  \textbf{44.6}       \\
\bottomrule
\end{tabular}
\caption{Comparisons of different kinds of combination for \jcfo{the} main modality and \jcfo{the} auxiliary modality in our cross-modal consensus module. \jcone{"Global" means that a convolution layer after global pooling is adopted to capture modal-specific global context, while "Local" means a convolution layer without global pooling but local-focused. }}
\label{tab:aggregation_type}
\vspace{-0.4cm}
\end{table}

\noindent\textbf{Effect on different kinds of combination for two modalities.} In our proposed CCM, we treat one modality as the main modality and another as \jcfo{the} auxiliary modality. In Section \ref{sec:CCM}, we utilize the main modality to generate the modality-specific global-aware descriptor $M^G$, while the auxiliary modality derives the cross-modal local-focused descriptor $M^L$ via a convolution layer. Here, we evaluate the different kinds of combination \ftthree{of} \jcfo{the} main and auxiliary modality in our cross-modal consensus module. The results are reported in Table \ref{tab:aggregation_type}. We can find that obtaining the global context information from the main modality or auxiliary can obtain \ftthree{stable} improvement compared with the results of the first row in Table \ref{tab:aggregation_type}. \eg the results of third row outperforms that of the first row by 1.6\% in AVG result. It verifies that obtaining global context information benefit to guide the recognition of information redundancy. In addition, when we obtain the global context information from the main modality, we can get the best results. 
Because \ftone{we aims to remove the task-irrelevant information redundancy from the main modality instead of \jcfo{the} auxiliary modality, and obtaining the global context information from the main modality can 
handle the overall \ftthree{information} of the main modality.}


\begin{figure}[t]
    \centering
    \includegraphics[width=\linewidth]{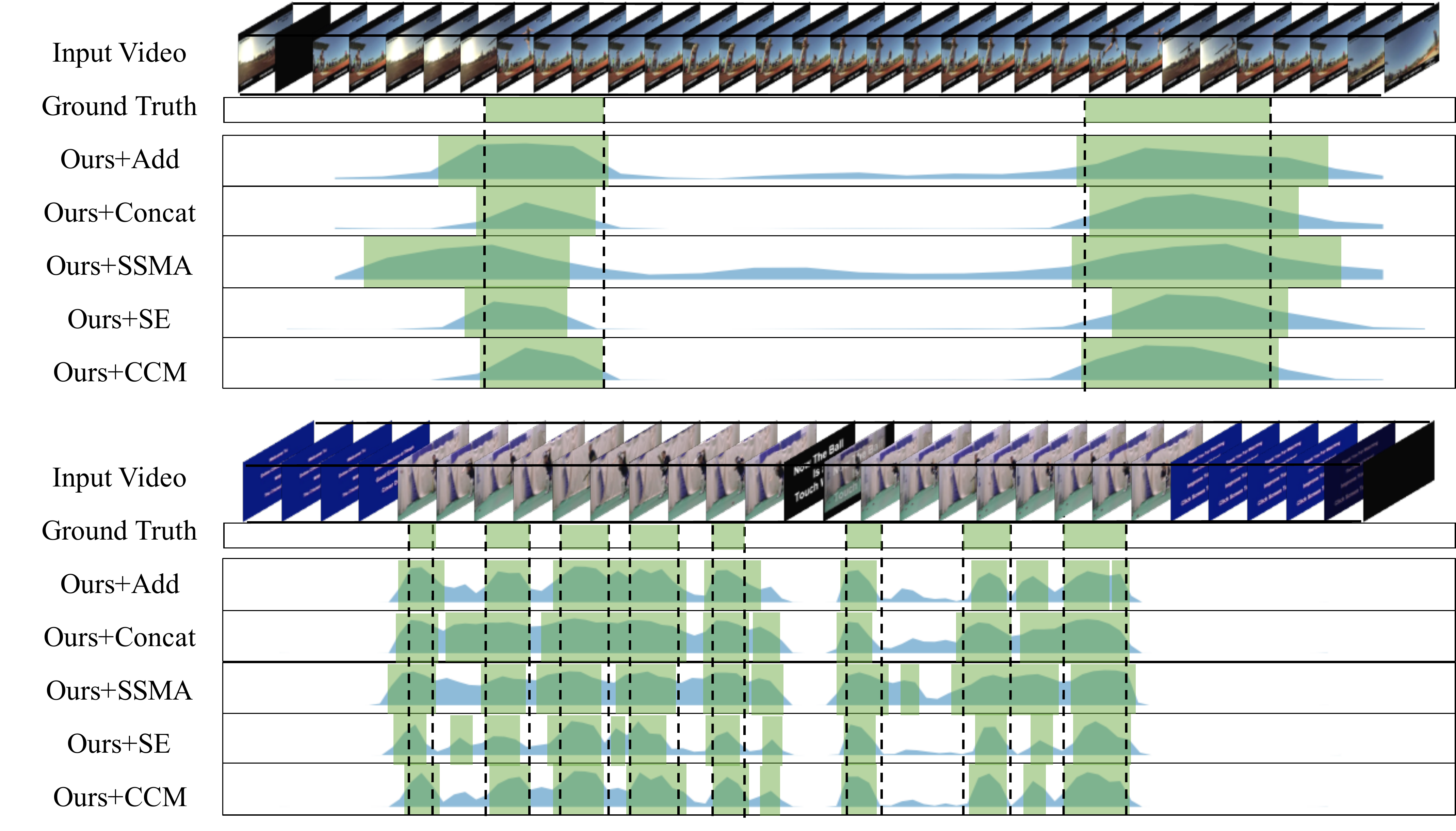}
    \caption{The illustration of the action localization results predicted by our full method and several variant methods on several video samples. Action proposals are represented by green boxes. The horizontal and vertical axes are time and intensity of attention, respectively. \ftthree{The method ``Ours + CCM'' means our full method CO$_2$-Net.}}
    \vspace{-5pt}
    \label{fig:visual}
\end{figure}
\begin{table}[t]
    \centering
    \begin{tabular}{c|c|c|c|c|c}
    \toprule
    method               &   Add    & Concat    & SSMA \cite{valada2019self} & SE \cite{hu2018squeeze}    & CCM \\
    \midrule
        Avg mAP             &   39.9    &  39.5   &  38.0  & 43.0 &  \textbf{44.6}  \\
    \bottomrule
    \end{tabular}
    \caption{Comparisons with other multi-modal early fusion methods \ftthree{(\ie addition and concatenation), SSMA \cite{valada2019self} and} SE-attention \cite{hu2018squeeze} in CO$_2$-Net in term of average mAP under multiple IoU thresholds \{0.1:0.7:0.1\}.}
    \label{tab:fusion_type}
    \vspace{-0.6cm}
\end{table}

\noindent\textbf{Compare with other fusion methods.} To verify that our proposed cross-modal consensus module is more suitable than other fusion methods for WS-TAL, we compare with other fusion methods and report results in Table \ref{tab:fusion_type}, in which SSMA is the fusion method in \cite{valada2019self}. 
We can find that our proposed CCM gains the best results compared with other fusion methods, \eg our CO$_2$-Net with CCM outperforms that with ``Concate'' by 5.1\%. We can also find that the SSMA even underperforms the ``Add'' and ``Concat'', because it contains a specific structure that does not suitable for temporal action localization. \ftthree{The SE-attention mechanism \cite{hu2018squeeze} can also gain an improvement compared with those early fusion methods, but results of our method still outperform that of \ftfour{SE} by 1.6\%.} The results in Table \ref{tab:fusion_type} verify that our proposed cross-modal consensus module can better fuse two modalities to boost the performance than those fusion \ftthree{methods}. Moreover, \jctwo{Though CO$_2$-Net also concatenate two types of features after filtering information redundancy with CCM, \ftthree{the results of ``Concat'' and ``CCM'' shown in Table \ref{tab:fusion_type} indicate that our method with CCM performs much better than the method with "Concat" on the original features}, showing the significance of feature re-calibration for more representative features.}

\vspace{-0.1cm}
\subsection{Visual Results}
To better illustrate our method, we also illustrate the detected results of several samples in Figure \ref{fig:visual} using the methods in Table \ref{tab:fusion_type}. It is obvious that our method  with CCM \jctwo{can predict more accurate localization against than the other fusion methods, showing the significance of removing the task-irrelevant information redundancy and the efficacy of our CCM.}

\vspace{-0.1cm}
\section{Conclusion}
In this work, \jcfi{we explore feature re-calibration for action localization to reduce the redundancy.} A cross-modal consensus network \jcfi{is proposed to tackle this problem}. 
\fttwo{We utilize a cross-modal consensus module to filter out the information redundancy in the main modality with the help of information from different perspectives of \jcfo{the} auxiliary modality.}
Also, we apply a mutual learning loss to enable \jctwo{two \jcth{cross-modal consensus modules} to learn from each other for mutual promotion.} Finally, we conduct extensive experiments to verify the effectiveness of our CO$_2$-Net and the results on ablation studies show that our proposed cross-modal consensus module can help to produce more representative features that would boost the performance of WS-TAL.

\jcfi{
\section{Acknowledgement}
This work was supported partially by the NSFC (U1911401, U1811461), Guangdong NSF Project (No.2020B1515120085, 2018B030312002), the Key-Area Research and Development Program of Guangzhou (202007030004), the Early Career Scheme of the Research Grants Council (RGC) of the Hong Kong SAR under grant No.~26202321 and HKUST Startup Fund No.~R9253. Work was partilly done during Fa-Ting's and Jia-Chang's internship in ARC, PCG Tencent.
}

\bibliographystyle{ACM-Reference-Format}
\bibliography{reference}


\begin{thebibliography}{54}


\ifx \showCODEN    \undefined \def \showCODEN     #1{\unskip}     \fi
\ifx \showDOI      \undefined \def \showDOI       #1{#1}\fi
\ifx \showISBNx    \undefined \def \showISBNx     #1{\unskip}     \fi
\ifx \showISBNxiii \undefined \def \showISBNxiii  #1{\unskip}     \fi
\ifx \showISSN     \undefined \def \showISSN      #1{\unskip}     \fi
\ifx \showLCCN     \undefined \def \showLCCN      #1{\unskip}     \fi
\ifx \shownote     \undefined \def \shownote      #1{#1}          \fi
\ifx \showarticletitle \undefined \def \showarticletitle #1{#1}   \fi
\ifx \showURL      \undefined \def \showURL       {\relax}        \fi
\providecommand\bibfield[2]{#2}
\providecommand\bibinfo[2]{#2}
\providecommand\natexlab[1]{#1}
\providecommand\showeprint[2][]{arXiv:#2}

\bibitem[\protect\citeauthoryear{Afouras, Owens, Chung, and Zisserman}{Afouras
  et~al\mbox{.}}{2020}]%
        {afouras2020self}
\bibfield{author}{\bibinfo{person}{Triantafyllos Afouras},
  \bibinfo{person}{Andrew Owens}, \bibinfo{person}{Joon~Son Chung}, {and}
  \bibinfo{person}{Andrew Zisserman}.} \bibinfo{year}{2020}\natexlab{}.
\newblock \showarticletitle{Self-supervised learning of audio-visual objects
  from video}.
\newblock \bibinfo{journal}{\emph{arXiv}} (\bibinfo{year}{2020}).
\newblock


\bibitem[\protect\citeauthoryear{Alwassel, Giancola, and Ghanem}{Alwassel
  et~al\mbox{.}}{2020}]%
        {alwassel2020tsp}
\bibfield{author}{\bibinfo{person}{Humam Alwassel}, \bibinfo{person}{Silvio
  Giancola}, {and} \bibinfo{person}{Bernard Ghanem}.}
  \bibinfo{year}{2020}\natexlab{}.
\newblock \showarticletitle{Tsp: Temporally-sensitive pretraining of video
  encoders for localization tasks}.
\newblock \bibinfo{journal}{\emph{arXiv preprint arXiv:2011.11479}}
  (\bibinfo{year}{2020}).
\newblock


\bibitem[\protect\citeauthoryear{Carreira and Zisserman}{Carreira and
  Zisserman}{2017}]%
        {carreira2017quo}
\bibfield{author}{\bibinfo{person}{Joao Carreira} {and} \bibinfo{person}{Andrew
  Zisserman}.} \bibinfo{year}{2017}\natexlab{}.
\newblock \showarticletitle{Quo vadis, action recognition? a new model and the
  kinetics dataset}. In \bibinfo{booktitle}{\emph{CVPR}}.
\newblock


\bibitem[\protect\citeauthoryear{Chao, Vijayanarasimhan, Seybold, Ross, Deng,
  and Sukthankar}{Chao et~al\mbox{.}}{2018}]%
        {chao2018rethinking}
\bibfield{author}{\bibinfo{person}{Yu-Wei Chao}, \bibinfo{person}{Sudheendra
  Vijayanarasimhan}, \bibinfo{person}{Bryan Seybold}, \bibinfo{person}{David~A
  Ross}, \bibinfo{person}{Jia Deng}, {and} \bibinfo{person}{Rahul Sukthankar}.}
  \bibinfo{year}{2018}\natexlab{}.
\newblock \showarticletitle{Rethinking the faster r-cnn architecture for
  temporal action localization}. In \bibinfo{booktitle}{\emph{CVPR}}.
\newblock


\bibitem[\protect\citeauthoryear{Choe and Shim}{Choe and Shim}{2019}]%
        {choe2019attention}
\bibfield{author}{\bibinfo{person}{Junsuk Choe} {and} \bibinfo{person}{Hyunjung
  Shim}.} \bibinfo{year}{2019}\natexlab{}.
\newblock \showarticletitle{Attention-based dropout layer for weakly supervised
  object localization}. In \bibinfo{booktitle}{\emph{CVPR}}.
\newblock


\bibitem[\protect\citeauthoryear{Deng, Chen, Liu, Gao, and Tao}{Deng
  et~al\mbox{.}}{2018}]%
        {deng2018triplet}
\bibfield{author}{\bibinfo{person}{Cheng Deng}, \bibinfo{person}{Zhaojia Chen},
  \bibinfo{person}{Xianglong Liu}, \bibinfo{person}{Xinbo Gao}, {and}
  \bibinfo{person}{Dacheng Tao}.} \bibinfo{year}{2018}\natexlab{}.
\newblock \showarticletitle{Triplet-based deep hashing network for cross-modal
  retrieval}.
\newblock \bibinfo{journal}{\emph{TIP}} (\bibinfo{year}{2018}).
\newblock


\bibitem[\protect\citeauthoryear{Fabian Caba~Heilbron and Niebles}{Fabian
  Caba~Heilbron and Niebles}{2015}]%
        {caba2015activitynet}
\bibfield{author}{\bibinfo{person}{Bernard~Ghanem Fabian Caba~Heilbron,
  Victor~Escorcia} {and} \bibinfo{person}{Juan~Carlos Niebles}.}
  \bibinfo{year}{2015}\natexlab{}.
\newblock \showarticletitle{ActivityNet: A Large-Scale Video Benchmark for
  Human Activity Understanding}. In \bibinfo{booktitle}{\emph{CVPR}}.
\newblock


\bibitem[\protect\citeauthoryear{Feng, Hong, and Zheng}{Feng
  et~al\mbox{.}}{2021}]%
        {feng2021mist}
\bibfield{author}{\bibinfo{person}{Jia-Chang Feng}, \bibinfo{person}{Fa-Ting
  Hong}, {and} \bibinfo{person}{Wei-Shi Zheng}.}
  \bibinfo{year}{2021}\natexlab{}.
\newblock \showarticletitle{MIST: Multiple Instance Self-Training Framework for
  Video Anomaly Detection}. In \bibinfo{booktitle}{\emph{CVPR}}.
\newblock


\bibitem[\protect\citeauthoryear{Gong, Wang, Mu, and Tian}{Gong
  et~al\mbox{.}}{2020}]%
        {gong2020learning}
\bibfield{author}{\bibinfo{person}{Guoqiang Gong}, \bibinfo{person}{Xinghan
  Wang}, \bibinfo{person}{Yadong Mu}, {and} \bibinfo{person}{Qi Tian}.}
  \bibinfo{year}{2020}\natexlab{}.
\newblock \showarticletitle{Learning Temporal Co-Attention Models for
  Unsupervised Video Action Localization}. In \bibinfo{booktitle}{\emph{CVPR}}.
\newblock


\bibitem[\protect\citeauthoryear{Hong, Huang, Li, and Zheng}{Hong
  et~al\mbox{.}}{2020}]%
        {hong2020mini}
\bibfield{author}{\bibinfo{person}{Fa-Ting Hong}, \bibinfo{person}{Xuanteng
  Huang}, \bibinfo{person}{Wei-Hong Li}, {and} \bibinfo{person}{Wei-Shi
  Zheng}.} \bibinfo{year}{2020}\natexlab{}.
\newblock \showarticletitle{MINI-Net: Multiple Instance Ranking Network for
  Video Highlight Detection}. In \bibinfo{booktitle}{\emph{ECCV}}.
\newblock


\bibitem[\protect\citeauthoryear{Hu, Shen, and Sun}{Hu et~al\mbox{.}}{2018}]%
        {hu2018squeeze}
\bibfield{author}{\bibinfo{person}{Jie Hu}, \bibinfo{person}{Li Shen}, {and}
  \bibinfo{person}{Gang Sun}.} \bibinfo{year}{2018}\natexlab{}.
\newblock \showarticletitle{Squeeze-and-excitation networks}. In
  \bibinfo{booktitle}{\emph{CVPR}}.
\newblock


\bibitem[\protect\citeauthoryear{Islam, Long, and Radke}{Islam
  et~al\mbox{.}}{2021}]%
        {islam2021hybrid}
\bibfield{author}{\bibinfo{person}{Ashraful Islam}, \bibinfo{person}{Chengjiang
  Long}, {and} \bibinfo{person}{Richard~J Radke}.}
  \bibinfo{year}{2021}\natexlab{}.
\newblock \showarticletitle{A Hybrid Attention Mechanism for Weakly-Supervised
  Temporal Action Localization}.
\newblock \bibinfo{journal}{\emph{arXiv}} (\bibinfo{year}{2021}).
\newblock


\bibitem[\protect\citeauthoryear{Islam and Radke}{Islam and Radke}{2020}]%
        {islam2020weakly}
\bibfield{author}{\bibinfo{person}{Ashraful Islam} {and}
  \bibinfo{person}{Richard Radke}.} \bibinfo{year}{2020}\natexlab{}.
\newblock \showarticletitle{Weakly Supervised Temporal Action Localization
  Using Deep Metric Learning}. In \bibinfo{booktitle}{\emph{WACV}}.
\newblock


\bibitem[\protect\citeauthoryear{Jain, Ghodrati, and Snoek}{Jain
  et~al\mbox{.}}{2020}]%
        {jain2020actionbytes}
\bibfield{author}{\bibinfo{person}{Mihir Jain}, \bibinfo{person}{Amir
  Ghodrati}, {and} \bibinfo{person}{Cees~GM Snoek}.}
  \bibinfo{year}{2020}\natexlab{}.
\newblock \showarticletitle{ActionBytes: Learning from trimmed videos to
  localize actions}. In \bibinfo{booktitle}{\emph{CVPR}}.
\newblock


\bibitem[\protect\citeauthoryear{Jiang, Liu, Roshan~Zamir, Toderici, Laptev,
  Shah, and Sukthankar}{Jiang et~al\mbox{.}}{2014}]%
        {THUMOS14}
\bibfield{author}{\bibinfo{person}{Y.-G. Jiang}, \bibinfo{person}{J. Liu},
  \bibinfo{person}{A. Roshan~Zamir}, \bibinfo{person}{G. Toderici},
  \bibinfo{person}{I. Laptev}, \bibinfo{person}{M. Shah}, {and}
  \bibinfo{person}{R. Sukthankar}.} \bibinfo{year}{2014}\natexlab{}.
\newblock \bibinfo{title}{{THUMOS} Challenge: Action Recognition with a Large
  Number of Classes}.
\newblock \bibinfo{howpublished}{\url{http://crcv.ucf.edu/THUMOS14/}}.
\newblock


\bibitem[\protect\citeauthoryear{Jing, Wang, Wang, and Tan}{Jing
  et~al\mbox{.}}{2020}]%
        {jing2020cross}
\bibfield{author}{\bibinfo{person}{Ya Jing}, \bibinfo{person}{Wei Wang},
  \bibinfo{person}{Liang Wang}, {and} \bibinfo{person}{Tieniu Tan}.}
  \bibinfo{year}{2020}\natexlab{}.
\newblock \showarticletitle{Cross-Modal Cross-Domain Moment Alignment Network
  for Person Search}. In \bibinfo{booktitle}{\emph{CVPR}}.
\newblock


\bibitem[\protect\citeauthoryear{Kay, Carreira, Simonyan, Zhang, Hillier,
  Vijayanarasimhan, Viola, Green, Back, Natsev, et~al\mbox{.}}{Kay
  et~al\mbox{.}}{2017}]%
        {kay2017kinetics}
\bibfield{author}{\bibinfo{person}{Will Kay}, \bibinfo{person}{Joao Carreira},
  \bibinfo{person}{Karen Simonyan}, \bibinfo{person}{Brian Zhang},
  \bibinfo{person}{Chloe Hillier}, \bibinfo{person}{Sudheendra
  Vijayanarasimhan}, \bibinfo{person}{Fabio Viola}, \bibinfo{person}{Tim
  Green}, \bibinfo{person}{Trevor Back}, \bibinfo{person}{Paul Natsev},
  {et~al\mbox{.}}} \bibinfo{year}{2017}\natexlab{}.
\newblock \showarticletitle{The kinetics human action video dataset}.
\newblock \bibinfo{journal}{\emph{arXiv}} (\bibinfo{year}{2017}).
\newblock


\bibitem[\protect\citeauthoryear{Kingma and Ba}{Kingma and Ba}{2014}]%
        {kingma2014adam}
\bibfield{author}{\bibinfo{person}{Diederik~P Kingma} {and}
  \bibinfo{person}{Jimmy Ba}.} \bibinfo{year}{2014}\natexlab{}.
\newblock \showarticletitle{Adam: A method for stochastic optimization}.
\newblock \bibinfo{journal}{\emph{arXiv}} (\bibinfo{year}{2014}).
\newblock


\bibitem[\protect\citeauthoryear{Lee, Uh, and Byun}{Lee et~al\mbox{.}}{2020}]%
        {lee2020background}
\bibfield{author}{\bibinfo{person}{Pilhyeon Lee}, \bibinfo{person}{Youngjung
  Uh}, {and} \bibinfo{person}{Hyeran Byun}.} \bibinfo{year}{2020}\natexlab{}.
\newblock \showarticletitle{Background Suppression Network for
  Weakly-Supervised Temporal Action Localization.}. In
  \bibinfo{booktitle}{\emph{AAAI}}.
\newblock


\bibitem[\protect\citeauthoryear{Lee, Wang, Lu, and Byun}{Lee
  et~al\mbox{.}}{2021}]%
        {lee2021Weakly}
\bibfield{author}{\bibinfo{person}{Pilhyeon Lee}, \bibinfo{person}{Jinglu
  Wang}, \bibinfo{person}{Yan Lu}, {and} \bibinfo{person}{Hyeran Byun}.}
  \bibinfo{year}{2021}\natexlab{}.
\newblock \showarticletitle{Weakly-supervised Temporal Action Localization by
  Uncertainty Modeling}.
\newblock \bibinfo{journal}{\emph{arXiv}} (\bibinfo{year}{2021}).
\newblock


\bibitem[\protect\citeauthoryear{Lei, Li, Zhou, Gan, Berg, Bansal, and Liu}{Lei
  et~al\mbox{.}}{2021}]%
        {lei2021less}
\bibfield{author}{\bibinfo{person}{Jie Lei}, \bibinfo{person}{Linjie Li},
  \bibinfo{person}{Luowei Zhou}, \bibinfo{person}{Zhe Gan},
  \bibinfo{person}{Tamara~L Berg}, \bibinfo{person}{Mohit Bansal}, {and}
  \bibinfo{person}{Jingjing Liu}.} \bibinfo{year}{2021}\natexlab{}.
\newblock \showarticletitle{Less is more: Clipbert for video-and-language
  learning via sparse sampling}. In \bibinfo{booktitle}{\emph{Proceedings of
  the IEEE/CVF Conference on Computer Vision and Pattern Recognition}}.
  \bibinfo{pages}{7331--7341}.
\newblock


\bibitem[\protect\citeauthoryear{Lin, Zhao, Su, Wang, and Yang}{Lin
  et~al\mbox{.}}{2018}]%
        {lin2018bsn}
\bibfield{author}{\bibinfo{person}{Tianwei Lin}, \bibinfo{person}{Xu Zhao},
  \bibinfo{person}{Haisheng Su}, \bibinfo{person}{Chongjing Wang}, {and}
  \bibinfo{person}{Ming Yang}.} \bibinfo{year}{2018}\natexlab{}.
\newblock \showarticletitle{Bsn: Boundary sensitive network for temporal action
  proposal generation}. In \bibinfo{booktitle}{\emph{ECCV}}.
\newblock


\bibitem[\protect\citeauthoryear{Liu, Jiang, and Wang}{Liu
  et~al\mbox{.}}{2019}]%
        {liu2019completeness}
\bibfield{author}{\bibinfo{person}{Daochang Liu}, \bibinfo{person}{Tingting
  Jiang}, {and} \bibinfo{person}{Yizhou Wang}.}
  \bibinfo{year}{2019}\natexlab{}.
\newblock \showarticletitle{Completeness modeling and context separation for
  weakly supervised temporal action localization}. In
  \bibinfo{booktitle}{\emph{CVPR}}.
\newblock


\bibitem[\protect\citeauthoryear{Liu, Wang, Zhang, Tang, Yuan, Nanning, and
  Hua}{Liu et~al\mbox{.}}{2021}]%
        {liu2021acsnet}
\bibfield{author}{\bibinfo{person}{Ziyi Liu}, \bibinfo{person}{Le Wang},
  \bibinfo{person}{Qilin Zhang}, \bibinfo{person}{Wei Tang},
  \bibinfo{person}{Junsong Yuan}, \bibinfo{person}{Zheng Nanning}, {and}
  \bibinfo{person}{Gang Hua}.} \bibinfo{year}{2021}\natexlab{}.
\newblock \showarticletitle{ACSNet: Action-Context Separation Network for
  Weakly Supervised Temporal Action Localization}. In
  \bibinfo{booktitle}{\emph{AAAI}}.
\newblock


\bibitem[\protect\citeauthoryear{Luo, Guillory, Shi, Ke, Wan, Darrell, and
  Xu}{Luo et~al\mbox{.}}{2020}]%
        {luo2020weakly}
\bibfield{author}{\bibinfo{person}{Zhekun Luo}, \bibinfo{person}{Devin
  Guillory}, \bibinfo{person}{Baifeng Shi}, \bibinfo{person}{Wei Ke},
  \bibinfo{person}{Fang Wan}, \bibinfo{person}{Trevor Darrell}, {and}
  \bibinfo{person}{Huijuan Xu}.} \bibinfo{year}{2020}\natexlab{}.
\newblock \showarticletitle{Weakly-Supervised Action Localization with
  Expectation-Maximization Multi-Instance Learning}.
\newblock \bibinfo{journal}{\emph{arXiv}} (\bibinfo{year}{2020}).
\newblock


\bibitem[\protect\citeauthoryear{Ma, Zhu, Yang, Zha, Kundu, Feiszli, and
  Shou}{Ma et~al\mbox{.}}{2020}]%
        {ma2020sf}
\bibfield{author}{\bibinfo{person}{Fan Ma}, \bibinfo{person}{Linchao Zhu},
  \bibinfo{person}{Yi Yang}, \bibinfo{person}{Shengxin Zha},
  \bibinfo{person}{Gourab Kundu}, \bibinfo{person}{Matt Feiszli}, {and}
  \bibinfo{person}{Zheng Shou}.} \bibinfo{year}{2020}\natexlab{}.
\newblock \showarticletitle{SF-Net: Single-frame supervision for temporal
  action localization}. In \bibinfo{booktitle}{\emph{ECCV}}.
\newblock


\bibitem[\protect\citeauthoryear{Min and Corso}{Min and Corso}{2020}]%
        {min2020adversarial}
\bibfield{author}{\bibinfo{person}{Kyle Min} {and} \bibinfo{person}{Jason~J
  Corso}.} \bibinfo{year}{2020}\natexlab{}.
\newblock \showarticletitle{Adversarial background-aware loss for
  weakly-supervised temporal activity localization}. In
  \bibinfo{booktitle}{\emph{ECCV}}.
\newblock


\bibitem[\protect\citeauthoryear{Munro and Damen}{Munro and Damen}{2020}]%
        {munro2020multi}
\bibfield{author}{\bibinfo{person}{Jonathan Munro} {and} \bibinfo{person}{Dima
  Damen}.} \bibinfo{year}{2020}\natexlab{}.
\newblock \showarticletitle{Multi-Modal Domain Adaptation for Fine-Grained
  Action Recognition}. In \bibinfo{booktitle}{\emph{CVPR}}.
\newblock


\bibitem[\protect\citeauthoryear{Narayan, Cholakkal, Khan, and Shao}{Narayan
  et~al\mbox{.}}{2019}]%
        {narayan20193c}
\bibfield{author}{\bibinfo{person}{Sanath Narayan}, \bibinfo{person}{Hisham
  Cholakkal}, \bibinfo{person}{Fahad~Shahbaz Khan}, {and} \bibinfo{person}{Ling
  Shao}.} \bibinfo{year}{2019}\natexlab{}.
\newblock \showarticletitle{3c-net: Category count and center loss for
  weakly-supervised action localization}. In \bibinfo{booktitle}{\emph{ICCV}}.
\newblock


\bibitem[\protect\citeauthoryear{Nawhal and Mori}{Nawhal and Mori}{2021}]%
        {nawhal2021activity}
\bibfield{author}{\bibinfo{person}{Megha Nawhal} {and} \bibinfo{person}{Greg
  Mori}.} \bibinfo{year}{2021}\natexlab{}.
\newblock \showarticletitle{Activity Graph Transformer for Temporal Action
  Localization}.
\newblock \bibinfo{journal}{\emph{arXiv}} (\bibinfo{year}{2021}).
\newblock


\bibitem[\protect\citeauthoryear{Ngiam, Khosla, Kim, Nam, Lee, and Ng}{Ngiam
  et~al\mbox{.}}{2011}]%
        {ngiam2011multimodal}
\bibfield{author}{\bibinfo{person}{Jiquan Ngiam}, \bibinfo{person}{Aditya
  Khosla}, \bibinfo{person}{Mingyu Kim}, \bibinfo{person}{Juhan Nam},
  \bibinfo{person}{Honglak Lee}, {and} \bibinfo{person}{Andrew~Y Ng}.}
  \bibinfo{year}{2011}\natexlab{}.
\newblock \showarticletitle{Multimodal deep learning}. In
  \bibinfo{booktitle}{\emph{ICML}}.
\newblock


\bibitem[\protect\citeauthoryear{Nguyen, Liu, Prasad, and Han}{Nguyen
  et~al\mbox{.}}{2018}]%
        {nguyen2018weakly}
\bibfield{author}{\bibinfo{person}{Phuc Nguyen}, \bibinfo{person}{Ting Liu},
  \bibinfo{person}{Gautam Prasad}, {and} \bibinfo{person}{Bohyung Han}.}
  \bibinfo{year}{2018}\natexlab{}.
\newblock \showarticletitle{Weakly supervised action localization by sparse
  temporal pooling network}. In \bibinfo{booktitle}{\emph{CVPR}}.
\newblock


\bibitem[\protect\citeauthoryear{Pardo, Alwassel, Caba, Thabet, and
  Ghanem}{Pardo et~al\mbox{.}}{2021}]%
        {pardo2021refineloc}
\bibfield{author}{\bibinfo{person}{Alejandro Pardo}, \bibinfo{person}{Humam
  Alwassel}, \bibinfo{person}{Fabian Caba}, \bibinfo{person}{Ali Thabet}, {and}
  \bibinfo{person}{Bernard Ghanem}.} \bibinfo{year}{2021}\natexlab{}.
\newblock \showarticletitle{Refineloc: Iterative refinement for
  weakly-supervised action localization}. In \bibinfo{booktitle}{\emph{WACV}}.
\newblock


\bibitem[\protect\citeauthoryear{Paszke, Gross, Massa, Lerer, Bradbury, Chanan,
  Killeen, Lin, Gimelshein, Antiga, et~al\mbox{.}}{Paszke
  et~al\mbox{.}}{2019}]%
        {paszke2019pytorch}
\bibfield{author}{\bibinfo{person}{Adam Paszke}, \bibinfo{person}{Sam Gross},
  \bibinfo{person}{Francisco Massa}, \bibinfo{person}{Adam Lerer},
  \bibinfo{person}{James Bradbury}, \bibinfo{person}{Gregory Chanan},
  \bibinfo{person}{Trevor Killeen}, \bibinfo{person}{Zeming Lin},
  \bibinfo{person}{Natalia Gimelshein}, \bibinfo{person}{Luca Antiga},
  {et~al\mbox{.}}} \bibinfo{year}{2019}\natexlab{}.
\newblock \showarticletitle{Pytorch: An imperative style, high-performance deep
  learning library}.
\newblock \bibinfo{journal}{\emph{arXiv}} (\bibinfo{year}{2019}).
\newblock


\bibitem[\protect\citeauthoryear{Paul, Roy, and Roy-Chowdhury}{Paul
  et~al\mbox{.}}{2018}]%
        {paul2018w}
\bibfield{author}{\bibinfo{person}{Sujoy Paul}, \bibinfo{person}{Sourya Roy},
  {and} \bibinfo{person}{Amit~K Roy-Chowdhury}.}
  \bibinfo{year}{2018}\natexlab{}.
\newblock \showarticletitle{W-talc: Weakly-supervised temporal activity
  localization and classification}. In \bibinfo{booktitle}{\emph{ECCV}}.
\newblock


\bibitem[\protect\citeauthoryear{Rao, Xu, Xiong, Xu, Huang, Zhou, and Lin}{Rao
  et~al\mbox{.}}{2020}]%
        {rao2020a}
\bibfield{author}{\bibinfo{person}{Anyi Rao}, \bibinfo{person}{Linning Xu},
  \bibinfo{person}{Yu Xiong}, \bibinfo{person}{Guodong Xu},
  \bibinfo{person}{Qingqiu Huang}, \bibinfo{person}{Bolei Zhou}, {and}
  \bibinfo{person}{Dahua Lin}.} \bibinfo{year}{2020}\natexlab{}.
\newblock \showarticletitle{A Local-to-Global Approach to Multi-Modal Movie
  Scene Segmentation}. In \bibinfo{booktitle}{\emph{CVPR}}.
\newblock


\bibitem[\protect\citeauthoryear{Shi, Dai, Mu, and Wang}{Shi
  et~al\mbox{.}}{2020}]%
        {shi2020weakly}
\bibfield{author}{\bibinfo{person}{Baifeng Shi}, \bibinfo{person}{Qi Dai},
  \bibinfo{person}{Yadong Mu}, {and} \bibinfo{person}{Jingdong Wang}.}
  \bibinfo{year}{2020}\natexlab{}.
\newblock \showarticletitle{Weakly-supervised action localization by generative
  attention modeling}. In \bibinfo{booktitle}{\emph{CVPR}}.
\newblock


\bibitem[\protect\citeauthoryear{Shou, Gao, Zhang, Miyazawa, and Chang}{Shou
  et~al\mbox{.}}{2018}]%
        {shou2018autoloc}
\bibfield{author}{\bibinfo{person}{Zheng Shou}, \bibinfo{person}{Hang Gao},
  \bibinfo{person}{Lei Zhang}, \bibinfo{person}{Kazuyuki Miyazawa}, {and}
  \bibinfo{person}{Shih-Fu Chang}.} \bibinfo{year}{2018}\natexlab{}.
\newblock \showarticletitle{Autoloc: Weakly-supervised temporal action
  localization in untrimmed videos}. In \bibinfo{booktitle}{\emph{ECCV}}.
\newblock


\bibitem[\protect\citeauthoryear{Shou, Wang, and Chang}{Shou
  et~al\mbox{.}}{2016}]%
        {shou2016temporal}
\bibfield{author}{\bibinfo{person}{Zheng Shou}, \bibinfo{person}{Dongang Wang},
  {and} \bibinfo{person}{Shih-Fu Chang}.} \bibinfo{year}{2016}\natexlab{}.
\newblock \showarticletitle{Temporal action localization in untrimmed videos
  via multi-stage cnns}. In \bibinfo{booktitle}{\emph{CVPR}}.
\newblock


\bibitem[\protect\citeauthoryear{Sultani, Chen, and Shah}{Sultani
  et~al\mbox{.}}{2018}]%
        {sultani2018real}
\bibfield{author}{\bibinfo{person}{Waqas Sultani}, \bibinfo{person}{Chen Chen},
  {and} \bibinfo{person}{Mubarak Shah}.} \bibinfo{year}{2018}\natexlab{}.
\newblock \showarticletitle{Real-world anomaly detection in surveillance
  videos}. In \bibinfo{booktitle}{\emph{Proceedings of the IEEE conference on
  computer vision and pattern recognition}}. \bibinfo{pages}{6479--6488}.
\newblock


\bibitem[\protect\citeauthoryear{Valada, Mohan, and Burgard}{Valada
  et~al\mbox{.}}{2019}]%
        {valada2019self}
\bibfield{author}{\bibinfo{person}{Abhinav Valada}, \bibinfo{person}{Rohit
  Mohan}, {and} \bibinfo{person}{Wolfram Burgard}.}
  \bibinfo{year}{2019}\natexlab{}.
\newblock \showarticletitle{Self-supervised model adaptation for multimodal
  semantic segmentation}.
\newblock \bibinfo{journal}{\emph{IJCV}} (\bibinfo{year}{2019}).
\newblock


\bibitem[\protect\citeauthoryear{Vaswani, Shazeer, Parmar, Uszkoreit, Jones,
  Gomez, Kaiser, and Polosukhin}{Vaswani et~al\mbox{.}}{2017}]%
        {vaswani2017attention}
\bibfield{author}{\bibinfo{person}{Ashish Vaswani}, \bibinfo{person}{Noam
  Shazeer}, \bibinfo{person}{Niki Parmar}, \bibinfo{person}{Jakob Uszkoreit},
  \bibinfo{person}{Llion Jones}, \bibinfo{person}{Aidan~N Gomez},
  \bibinfo{person}{Lukasz Kaiser}, {and} \bibinfo{person}{Illia Polosukhin}.}
  \bibinfo{year}{2017}\natexlab{}.
\newblock \showarticletitle{Attention is all you need}.
\newblock \bibinfo{journal}{\emph{arXiv}} (\bibinfo{year}{2017}).
\newblock


\bibitem[\protect\citeauthoryear{Wang, Cui, Ou, and Zhu}{Wang
  et~al\mbox{.}}{2015}]%
        {wang2015learning}
\bibfield{author}{\bibinfo{person}{Daixin Wang}, \bibinfo{person}{Peng Cui},
  \bibinfo{person}{Mingdong Ou}, {and} \bibinfo{person}{Wenwu Zhu}.}
  \bibinfo{year}{2015}\natexlab{}.
\newblock \showarticletitle{Learning compact hash codes for multimodal
  representations using orthogonal deep structure}.
\newblock \bibinfo{journal}{\emph{IEEE Transactions on Multimedia}}
  (\bibinfo{year}{2015}).
\newblock


\bibitem[\protect\citeauthoryear{Wang, Hong, Tan, and Yuan}{Wang
  et~al\mbox{.}}{2019}]%
        {wang2019pruning}
\bibfield{author}{\bibinfo{person}{Zhenzhen Wang}, \bibinfo{person}{Weixiang
  Hong}, \bibinfo{person}{Yap-Peng Tan}, {and} \bibinfo{person}{Junsong Yuan}.}
  \bibinfo{year}{2019}\natexlab{}.
\newblock \showarticletitle{Pruning 3D Filters For Accelerating 3D ConvNets}.
\newblock \bibinfo{journal}{\emph{IEEE Transactions on Multimedia}}
  (\bibinfo{year}{2019}).
\newblock


\bibitem[\protect\citeauthoryear{Xu, Ouyang, Ricci, Wang, and Sebe}{Xu
  et~al\mbox{.}}{2017}]%
        {xu2017learning}
\bibfield{author}{\bibinfo{person}{Dan Xu}, \bibinfo{person}{Wanli Ouyang},
  \bibinfo{person}{Elisa Ricci}, \bibinfo{person}{Xiaogang Wang}, {and}
  \bibinfo{person}{Nicu Sebe}.} \bibinfo{year}{2017}\natexlab{}.
\newblock \showarticletitle{Learning Cross-Modal Deep Representations for
  Robust Pedestrian Detection}. In \bibinfo{booktitle}{\emph{CVPR}}.
\newblock


\bibitem[\protect\citeauthoryear{Xu, Ouyang, Wang, and Sebe}{Xu
  et~al\mbox{.}}{2018}]%
        {xu2018PAD-Net}
\bibfield{author}{\bibinfo{person}{Dan Xu}, \bibinfo{person}{Wanli Ouyang},
  \bibinfo{person}{Xiaogang Wang}, {and} \bibinfo{person}{Nicu Sebe}.}
  \bibinfo{year}{2018}\natexlab{}.
\newblock \showarticletitle{PAD-Net: Multi-Tasks Guided
  Prediciton-and-Distillation Network for Simultaneous Depth Estimation and
  Scene Parsing}. In \bibinfo{booktitle}{\emph{CVPR}}.
\newblock


\bibitem[\protect\citeauthoryear{Xu, Ricci, Yan, Song, and Sebe}{Xu
  et~al\mbox{.}}{2015}]%
        {xu2015learning}
\bibfield{author}{\bibinfo{person}{Dan Xu}, \bibinfo{person}{Elisa Ricci},
  \bibinfo{person}{Yan Yan}, \bibinfo{person}{Jingkuan Song}, {and}
  \bibinfo{person}{Nicu Sebe}.} \bibinfo{year}{2015}\natexlab{}.
\newblock \showarticletitle{Learning deep representations of appearance and
  motion for anomalous event detection}. In \bibinfo{booktitle}{\emph{BMVC}}.
\newblock


\bibitem[\protect\citeauthoryear{Xu, P{\'e}rez-R{\'u}a, Escorcia, Martinez,
  Zhu, Zhang, Ghanem, and Xiang}{Xu et~al\mbox{.}}{2020}]%
        {xu2020boundary}
\bibfield{author}{\bibinfo{person}{Mengmeng Xu}, \bibinfo{person}{Juan-Manuel
  P{\'e}rez-R{\'u}a}, \bibinfo{person}{Victor Escorcia}, \bibinfo{person}{Brais
  Martinez}, \bibinfo{person}{Xiatian Zhu}, \bibinfo{person}{Li Zhang},
  \bibinfo{person}{Bernard Ghanem}, {and} \bibinfo{person}{Tao Xiang}.}
  \bibinfo{year}{2020}\natexlab{}.
\newblock \showarticletitle{Boundary-sensitive pre-training for temporal
  localization in videos}.
\newblock \bibinfo{journal}{\emph{arXiv preprint arXiv:2011.10830}}
  (\bibinfo{year}{2020}).
\newblock


\bibitem[\protect\citeauthoryear{Xu, Zhang, Cheng, Xie, Niu, Pu, and Wu}{Xu
  et~al\mbox{.}}{2019}]%
        {xu2019segregated}
\bibfield{author}{\bibinfo{person}{Yunlu Xu}, \bibinfo{person}{Chengwei Zhang},
  \bibinfo{person}{Zhanzhan Cheng}, \bibinfo{person}{Jianwen Xie},
  \bibinfo{person}{Yi Niu}, \bibinfo{person}{Shiliang Pu}, {and}
  \bibinfo{person}{Fei Wu}.} \bibinfo{year}{2019}\natexlab{}.
\newblock \showarticletitle{Segregated temporal assembly recurrent networks for
  weakly supervised multiple action detection}. In
  \bibinfo{booktitle}{\emph{AAAI}}.
\newblock


\bibitem[\protect\citeauthoryear{Zeng, Hong, Zheng, Yu, Zeng, Wang, and
  Lai}{Zeng et~al\mbox{.}}{2020}]%
        {zeng2020hybrid}
\bibfield{author}{\bibinfo{person}{Ling-An Zeng}, \bibinfo{person}{Fa-Ting
  Hong}, \bibinfo{person}{Wei-Shi Zheng}, \bibinfo{person}{Qi-Zhi Yu},
  \bibinfo{person}{Wei Zeng}, \bibinfo{person}{Yao-Wei Wang}, {and}
  \bibinfo{person}{Jian-Huang Lai}.} \bibinfo{year}{2020}\natexlab{}.
\newblock \showarticletitle{Hybrid Dynamic-static Context-aware Attention
  Network for Action Assessment in Long Videos}. In
  \bibinfo{booktitle}{\emph{ACM MM}}.
\newblock


\bibitem[\protect\citeauthoryear{Zeng, Huang, Tan, Rong, Zhao, Huang, and
  Gan}{Zeng et~al\mbox{.}}{2019}]%
        {zeng2019graph}
\bibfield{author}{\bibinfo{person}{Runhao Zeng}, \bibinfo{person}{Wenbing
  Huang}, \bibinfo{person}{Mingkui Tan}, \bibinfo{person}{Yu Rong},
  \bibinfo{person}{Peilin Zhao}, \bibinfo{person}{Junzhou Huang}, {and}
  \bibinfo{person}{Chuang Gan}.} \bibinfo{year}{2019}\natexlab{}.
\newblock \showarticletitle{Graph convolutional networks for temporal action
  localization}. In \bibinfo{booktitle}{\emph{ICCV}}.
\newblock


\bibitem[\protect\citeauthoryear{Zhai, Wang, Tang, Zhang, Yuan, and Hua}{Zhai
  et~al\mbox{.}}{2020}]%
        {zhai2020two}
\bibfield{author}{\bibinfo{person}{Yuanhao Zhai}, \bibinfo{person}{Le Wang},
  \bibinfo{person}{Wei Tang}, \bibinfo{person}{Qilin Zhang},
  \bibinfo{person}{Junsong Yuan}, {and} \bibinfo{person}{Gang Hua}.}
  \bibinfo{year}{2020}\natexlab{}.
\newblock \showarticletitle{Two-stream consensus network for weakly-supervised
  temporal action localization}. In \bibinfo{booktitle}{\emph{ECCV}}.
\newblock


\bibitem[\protect\citeauthoryear{Zhang, Shi, Li, and Li}{Zhang
  et~al\mbox{.}}{2020}]%
        {zhang2020multi}
\bibfield{author}{\bibinfo{person}{Xiao-Yu Zhang}, \bibinfo{person}{Haichao
  Shi}, \bibinfo{person}{Changsheng Li}, {and} \bibinfo{person}{Peng Li}.}
  \bibinfo{year}{2020}\natexlab{}.
\newblock \showarticletitle{Multi-instance multi-label action recognition and
  localization based on spatio-temporal pre-trimming for untrimmed videos}. In
  \bibinfo{booktitle}{\emph{AAAI}}.
\newblock


\bibitem[\protect\citeauthoryear{Zhao, Xiong, Wang, Wu, Tang, and Lin}{Zhao
  et~al\mbox{.}}{2017}]%
        {zhao2017temporal}
\bibfield{author}{\bibinfo{person}{Yue Zhao}, \bibinfo{person}{Yuanjun Xiong},
  \bibinfo{person}{Limin Wang}, \bibinfo{person}{Zhirong Wu},
  \bibinfo{person}{Xiaoou Tang}, {and} \bibinfo{person}{Dahua Lin}.}
  \bibinfo{year}{2017}\natexlab{}.
\newblock \showarticletitle{Temporal action detection with structured segment
  networks}. In \bibinfo{booktitle}{\emph{ICCV}}.
\newblock


\end{thebibliography}

\end{document}